\documentclass[%
 reprint,
 amsmath,amssymb,
 aip,jcp
]{revtex4-2}

\usepackage{microtype}
\usepackage{graphicx}
\usepackage{caption}
\usepackage{subcaption}
\usepackage{booktabs} % for professional tables
\usepackage{multirow}
\usepackage{tabularx}

\usepackage{amsmath,amssymb}
\usepackage{mathtools}
\usepackage{bm}

\usepackage{hyperref}

\newcommand{\painn}{\textsc{PaiNN}}

\newcommand{\icm}{cm$^{-1}$}

\newcommand{\rr}{\vec{r}}
\newcommand{\q}{\mathbf{s}}

\newcommand{\mmu}{\vec{\mathbf{v}}}

\newcommand{\new}[1]{#1}

\usepackage{ragged2e}
\DeclareCaptionJustification{justified}{\justifying}
\begin{document}

\title{Equivariant message passing for the prediction of tensorial properties and molecular spectra}

\author{Kristof T. Sch\"utt}
\affiliation{Machine Learning Group, Technische Universit\"at Berlin, 10587 Berlin, Germany}
\affiliation{Berlin Institute for the Foundations of Learning and Data, 10587 Berlin, Germany}
\email{kristof.schuett@tu-berlin.de}
\author{Oliver T. Unke}
\affiliation{Machine Learning Group, Technische Universit\"at Berlin, 10587 Berlin, Germany}
\affiliation{Berlin Institute for the Foundations of Learning and Data, 10587 Berlin, Germany}
\author{Michael Gastegger}
\affiliation{Machine Learning Group, Technische Universit\"at Berlin, 10587 Berlin, Germany}
\affiliation{BASLEARN -- TU Berlin/BASF Joint Lab for Machine Learning, 10587 Berlin, Germany}
%\email{michael.gastegger@tu-berlin.de}

\date{\today}

\begin{abstract}
Message passing neural networks have become a method of choice for learning on graphs, in particular the prediction of chemical properties and the acceleration of molecular dynamics studies.
While they readily scale to large training data sets, previous approaches have proven to be less data efficient than kernel methods.
We identify limitations of invariant representations as a major reason and extend the message passing formulation to rotationally equivariant representations.
On this basis, we propose the \emph{polarizable atom interaction neural network} (\painn{}) and improve on common molecule benchmarks over previous networks, while reducing model size and inference time.
We leverage the equivariant atomwise representations obtained by \painn{} for the prediction of tensorial properties.
Finally, we apply this to the simulation of molecular spectra, achieving speedups of 4-5 orders of magnitude compared to the electronic structure reference.
\end{abstract}

\maketitle

\section{Introduction}
\label{sec:introduction}

Studying dynamics of chemical systems allows insight into processes such as reactions or the folding of proteins, and constitutes a fundamental challenge in computational chemistry.
Since the motion of atoms is governed by the laws of quantum mechanics, accurate \textit{ab initio} molecular dynamics (MD) simulations may require solving the Schrödinger equation for millions of time steps.
While the exact solution is infeasible to compute for all but the smallest systems, even fast approximations such as density functional theory quickly become prohibitive for large systems and the prediction of accurate spectra.
%even fast approximations such as density functional theory scale with $\mathcal{O}(n^3)$ with the number of electrons.
% On the other hand, fast classical force fields require only microseconds per evaluation, but suffer from lack of accuracy and are restricted to certain classes of systems.

Recently, machine learning potentials~\cite{behler2016perspective,Unke2020mlforce,von2020exploring} have gained popularity for studying systems ranging from small molecules at high levels of theory~\cite{chmiela2018,westermayr2020combining} to systems with thousands or millions of atoms~\cite{morawietz2016van,bartok2018machine,lu202086}.
In particular, message-passing neural networks~\cite{gilmer2017neural} (MPNNs) yield accurate predictions for chemical properties across chemical compound space and can handle large amounts of training data.
Albeit MPNNs have significantly increased in accuracy over the years (as well as in computational cost), kernel methods with manually crafted features \cite{chmiela2017machine,christensen2020fchl,bartok2010gaussian} have still proven to perform better when only small training sets are available.

While molecules are often represented as graphs, they are in fact interacting particles in a continuous 3d space.
Consequently, SchNet~\cite{schutt2017schnet} modeled message passes as continuous-filter convolutions over that space, albeit with rotationally invariant filters.
As \citet{miller2020relevance} pointed out, this leads to a loss of relevant directional, equivariant information.
\citet{klicpera2020directional} have introduced directional message-passing, the angular information here is restricted to the messages while the representation of nodes (atoms) remains rotationally invariant.
%Beyond that, explicitly including angles and increasing the size of the neural network leads to increased inference time which may be prohibitive for the computation of large MD trajectories.
While equivariant convolutions have been successfully applied in computer vision~\cite{cohen2016steerable,weiler2018learning,worrall2018cubenet}, previous approaches to molecular prediction~\cite{thomas2018tensor,anderson2019cormorant} have not reached the accuracy of their rotationally invariant counterparts.

In this work, we propose rotationally equivariant message passing and the \emph{polarizable atom interaction neural network (\painn{})} architecture as one instance of it.
We examine the limited capability of rotation-invariant representations to propagate directional information and show that equivariant representations do not suffer from this issue.
\painn{} outperforms invariant message passing networks on common molecular benchmarks and performs at small sample sizes on par with kernel methods that have been deemed to be more data-efficient than neural networks.
Beyond that, the rotationally equivariant representation of \painn{} enables the prediction of tensorial properties which we apply to the ring-polymer MD simulation of infrared and Raman spectra.
By an acceleration of 4-5 orders of magnitude, \painn{} makes the simulation of these spectra feasible, reducing the runtime in one case from projected 25 years to one hour.

\section{Related work}
\citet{behler2007generalized} introduced neural network potentials taking atom-centered symmetry functions based on distances and angles as features.
Graph neural networks~\cite{scarselli2008graph} for molecular graphs~\cite{duvenaud2015convolutional,kearnes2016molecular} and 3d geometries~\cite{schutt2017deep,gilmer2017neural,schutt2018schnet,unke2019physnet,lubbers2018hierarchical} do not require such manually crafted features, but learn embeddings of atom types and use graph convolutions or, more general, message passing~\cite{gilmer2017neural}, to model atom interactions based on interatomic distances.
\citet{klicpera2020directional} introduced directional message passing by including additional angular information in the message.

Steerable CNNs~\cite{cohen2016steerable,weiler20183d} allow to build equivariant filter banks according to known symmetries of associated feature types.
While these approaches work on grids, architectures such as Tensor Field Networks~\cite{thomas2018tensor}, Cormorant~\cite{anderson2019cormorant} and \textsc{NequIP}~\cite{batzner2021se} use equivariant convolutions based on spherical harmonics (SH) and Clebsch-Gordon (CG) transforms for point clouds.
\citet{kondor2018generalization} have described a general framework on equivariance and convolutions in neural networks focusing on irreducible representations.
In contrast, \painn{} models equivariant interactions in Cartesian space which is conceptually simpler and does not require tensor contractions with CG coefficients.
\new{
A similar approach was proposed by \citet{jing2021learning} with the GVP-GNN.
Both approaches are designed for distinct applications resulting in crucial differences in the design of the neural network architectures, most notably the message functions. The GVP-GNN has been designed for single-point protein sequence predictions, e.g., allowing the use of non-smooth components. In contrast, \painn{} is designed for simulations with millions of inference steps, requiring a fast message function and a smoothly differentiable model.}

\section{Equivariant message passing}\label{sec:eqmpnn}

\paragraph{Notation.} To clearly distinguish between the two concepts, we write feature vectors as $\mathbf{x} \in \mathbb{R}^{Fx1}$ and vectors in 3d coordinate space as $\vec{r} \in \mathbb{R}^{1x3}$. Vectorial features will be written as $\vec{\mathbf{x}} \in \mathbb{R}^{Fx3}$. Norms $\| \cdot \|$, scalar products $\langle \cdot, \cdot \rangle$ and tensor products $\otimes$ are calculated along the spatial dimension. All other operations are calculated along the feature dimension, if not stated otherwise. We write the Hadamard product as $\circ$.

\subsection{Message passing for 3d-embedded graphs}
MPNNs build complex representations of nodes within their local neighborhood in a graph through a repeated exchange of \emph{messages} followed by \emph{updates} of node features.
Here, we consider graphs embbedded in 3d Euclidean space, where edges are specified by the relative positions $\vec{r}_{ij} = \vec{r}_j - \vec{r}_i$ of nodes $i, j$ within a local neighborhood $\mathcal{N}(i) = \{ j \, | \, \|\vec{r}_{ij}\| \leq r_\text{cut} \}$.
In a chemistry context, this agrees with the fact that a large part of the energy variations can be attributed to local interactions, often conceptualized as bond lengths and angles.
Thus, only nodes within that range can interact directly, so the number of messages does not scale quadratically with the number of the nodes.
A general MPNN for embedded graphs can be written as
\begin{align}
\mathbf{m}_i^{t+1} &= \sum_{j \in \mathcal{N}(i)} \mathbf{M}_t(\mathbf{s}_i^t, \mathbf{s}_j^t, \vec{r}_{ij}) \\
\mathbf{s}_i^{t+1} &= \mathbf{U}_t \left( \mathbf{s}_i^{t},  \mathbf{m}_i^{t+1}  \right).
\label{eq:mp}
\end{align}
with the update and message functions $\mathbf{U}_t$ and $\mathbf{M}_t$, respectively~\cite{gilmer2017neural}.
Many neural network potentials and even conventional CNNs can be cast in this general framework.
\new{Rotational invariance of the representation can be ensured by choosing rotationally invariant message and update functions, which by definition need to fulfill}
\begin{align}
    \mathbf{f}(\vec{\mathbf{x}}) = \mathbf{f}(R \,\vec{\mathbf{x}}),
\end{align}
\new{for any rotation matrix $R \in \mathbb{R}^{3 \times 3}$.}

\subsection{Building equivariant MPNNs}
To obtain more expressive representations of local environments, neurons do not have to be scalar, but can be geometric objects such as vectors and tensors~\cite{hinton2011transforming,cohen2016steerable,thomas2018tensor,anderson2019cormorant}.
For the purpose of this work, we restrict ourselves to scalar and vectorial representations $\mathbf{s}_i^{t}$ and $\vec{\mathbf{v}}_i^{t}$, respectively, such that a corresponding message pass can be written as
\begin{align}
\vec{\mathbf{m}}_i^{v,t+1} &= \sum_{j \in \mathcal{N}(i)} \vec{\mathbf{M}}_t(\mathbf{s}_i^{t}, \mathbf{s}_j^{t}, \vec{\mathbf{v}}_i^{t}, \vec{\mathbf{v}}_j^{t}, \vec{r}_{ij})
\label{eq:mpv}
\end{align}
Message and update functions for scalar features can be defined analogously.
\new{Rotational equivariance of vector features $\vec{\mathbf{v}}_i^{t+1}$ can be ensured by employing rotationally equivariant functions $\vec{\mathbf{U}}_t$ and $\vec{\mathbf{M}}_t$, fulfilling}
\begin{align}
    R \, \vec{\mathbf{f}}(\vec{\mathbf{x}}) = \vec{\mathbf{f}}(R \,\vec{\mathbf{x}})
\end{align}
\new{for any rotation matrix $R \in \mathbb{R}^{3 \times 3}$, where the matrix-vector product is applied over the spatial dimension.}
This constitutes essentially a linearity constraint for directional information.
Therefore, any nonlinearities of the model need to be applied to scalar features.
In particular, equivariant MPNNs may use the following operations:
\begin{itemize}
    \item Any (nonlinear) function of scalars: $\mathbf{f}(\mathbf{s})$
    \item Scaling of vectors: $\mathbf{s} \circ \vec{\mathbf{v}}$
    \item Linear combinations of equivariant vectors: $\mathbf{W} \vec{\mathbf{v}}$
    \item Scalar products: $\mathbf{s} = \| \vec{\mathbf{v}} \|^2$, $\mathbf{s} = \langle \vec{\mathbf{v}}_1, \vec{\mathbf{v}}_2 \rangle$
    \item Vector products: $\vec{\mathbf{v}}_1 \times \vec{\mathbf{v}}_2$
\end{itemize}
Thus, directional information is preserved during message passing while invariant representations can be recovered from scalar products to obtain rotationally invariant predictions.

\subsection{Limits of rotationally invariant representations}
\begin{table}[tb]
\caption{Comparison of expressiveness and computational complexity of distances, angles and directions for a simple message function of the oxygen atom of a water molecule.}
\label{tab:equivariant}
\begin{center}
\begin{scriptsize}
\begin{tabularx}{\columnwidth}{>{\centering\arraybackslash}m{2cm}|>{\centering\arraybackslash}m{2.1cm}>{\centering\arraybackslash}m{2.1cm}>{\centering\arraybackslash}m{2.1cm}}
\toprule
\textbf{\textit{Features}} & \textbf{Distances} & \textbf{Angles} & \textbf{Directions} \\ \midrule
 \textit{\textbf{H$_2$O}} \vspace{0.5cm} & \includegraphics[height=1.5cm]{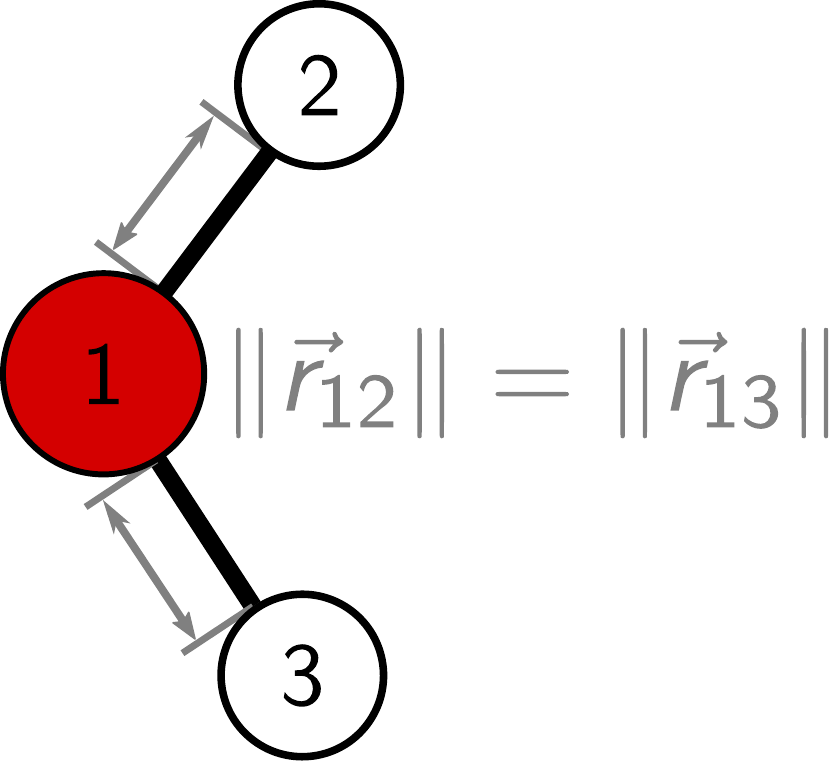} & \includegraphics[height=1.5cm]{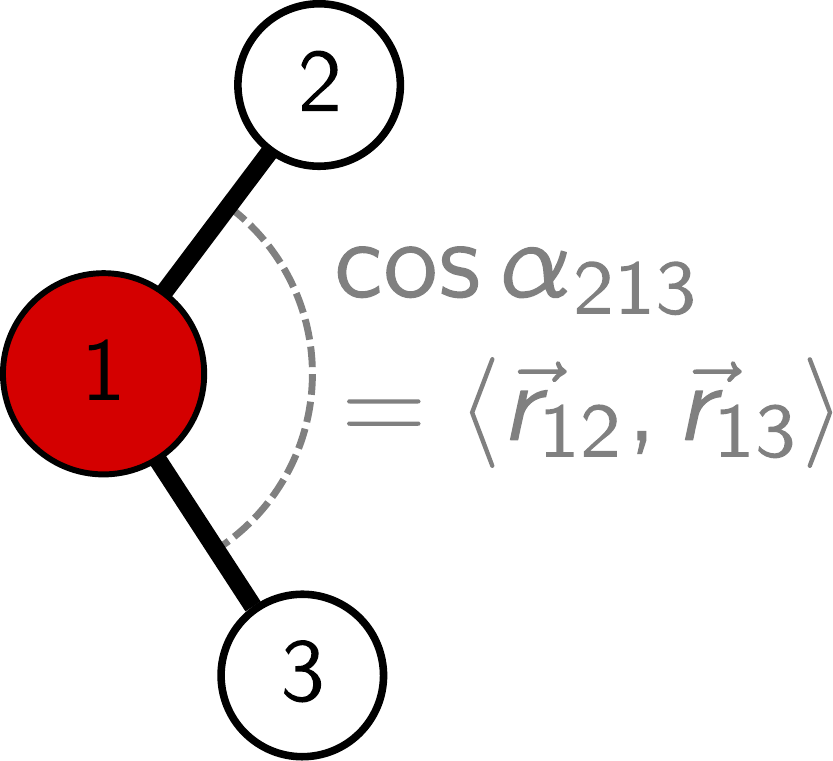} & \includegraphics[height=1.5cm]{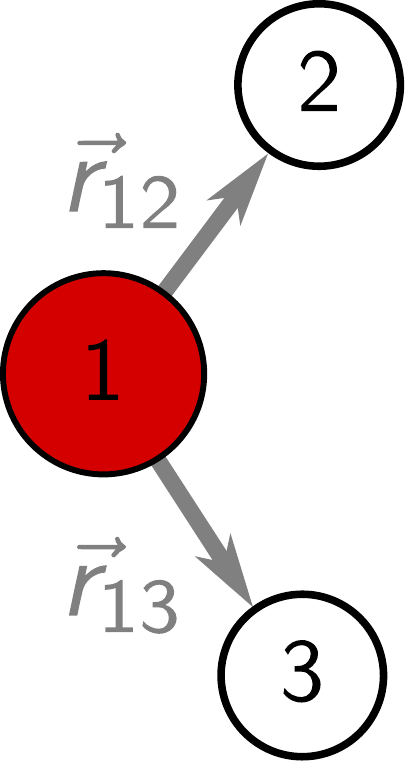}  \\
 \textit{\textbf{Message $M$ at atom $i$}} \vspace{0.3cm} & \[ \sum_{j \in \mathcal{N}_i} \| \vec{r}_{ij} \| \]  &  \[ \sum_{j \in \mathcal{N}_i} \sum_{k \in \mathcal{N}_{i}} \alpha_{jik} \]  &  \[ \sum_{j \in \mathcal{N}_i} \frac{\vec{r}_{ij}}{\| \vec{r}_{ij} \|}  \]  \\
\textit{\textbf{Scaling with neighbors}} & $\mathcal{O}(|\mathcal{N}|)$ \vspace{0.1cm} &  $\mathcal{O}(|\mathcal{N}|^2)$ \vspace{0.1cm} &  $\mathcal{O}(|\mathcal{N}|)$ \vspace{0.1cm} \\
\vspace{0.2cm} \textit{\textbf{Resolve change of $\| \vec{r}_{1j} \|$}} & yes \vspace{0.1cm} & no \vspace{0.1cm} &  no \vspace{0.1cm} \\
\vspace{0.2cm} \textit{\textbf{Resolve change of $\alpha_{213}$}} &   no \vspace{0.08cm}& yes \vspace{0.08cm}&  yes \vspace{0.08cm}\\
\bottomrule
\end{tabularx}
\end{scriptsize}
\end{center}
\end{table}
In the following, we examine the expressiveness of invariant and equivariant representations at the example of simple 2d molecular structures.
To ensure rotational invariance of a predicted property, the message function is often restricted to depend only on rotationally invariant inputs such as distances~\cite{schutt2017deep,lubbers2018hierarchical} or angles~\cite{klicpera2020directional,klicpera2020fast}.
While a single pass is limited to interactions within a local environment, successive message passes are supposed to construct more complex representations and propagate local information beyond the neighborhood.
This raises the question whether rotationally invariant representations of atomic environments $\mathbf{h}_i$ are sufficient here.

Tab.~\ref{tab:equivariant} compares three simplified message functions for the example of a water molecule.
Using a distance-based message function, the representation of the atom 1 (oxygen) is able to resolve changing bond lengths to atoms 2 and 3 (hydrogens), however it is not sensitive to changes of the bond angle.
On the other hand, using the angle directly as part of the message function can not resolve the distances.
Therefore, a combination of distances and angles is required to obtain a more expressive message function~\cite{klicpera2020directional}.
Unfortunately, including angles in the messages scales $\mathcal{O}(|\mathcal{N}|^2)$ with the number of neighbors.
Alternatively, directions to neighboring atoms may be used as messages $\vec{M}_t(\rr_{ij}) = \rr_{ij} / \|\rr_{ij}\|$.
Employing the update function $U_t(m) = \| m \|^2$, this is related to angles as follows:
\begin{align}
     \left\| \sum_{j=1}^N \frac{\rr_{ij}}{\| \rr_{ij}\|} \right\|^2
     = \sum_{j, k} \left\langle \frac{\rr_{ij}}{\| \rr_{ij}\|}, \frac{\rr_{ik}}{\| \rr_{ik}\|} \right\rangle 
     = \sum_{j=1}^N \sum_{k=1}^N \cos{\alpha_{jik}}. \notag
\end{align}
Thus, using equivariant messages, the runtime complexity remains $\mathcal{O}(|\mathcal{N}|)$ while angular information can be resolved.
Note that this update function contracts the equivariant messages to a rotationally invariant representation.

\begin{figure}[tb]
    \centering
    \includegraphics[width=\columnwidth]{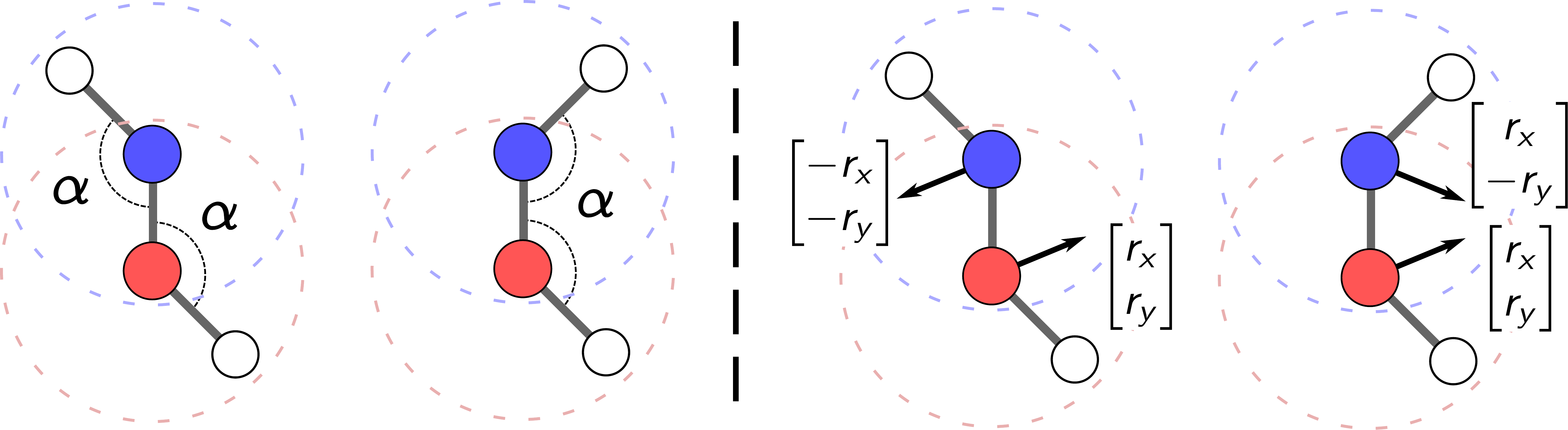}
    \caption{Illustration of message passing using angles and directions for two structures. All edges within the cutoff range (dashed lines) have equal length. The representations of the blue and red node are the same using angles (left), while directions allow to distinguish both structures (right). }
    \label{fig:dirprop}
\end{figure}

Beyond the computational benefits, equivariant representations allow to propagate directional information beyond the neighborhood which is \emph{not} possible in the invariant case.
Fig.~\ref{fig:dirprop} illustrates this at a minimal example, where four atoms are arranged in an equidistant chain with the cutoff chosen such that only neighboring atoms are within range.
We observe that for the two arrangements the angles are equal as well (Fig.~\ref{fig:dirprop}, left).
Therefore, they are indistinguishable for invariant message passing with distances and angles.
In contrast, the equivariant representations differ in sign of their components (Fig.~\ref{fig:dirprop}, right).
When contracting them to invariant representations, as in the previous example, this information is lost.
However, we may instead retain equivariance in the representation and design a message function $\vec{\mathbf{M}}_t$ that does not only propagate directions to neighboring atoms, but those of equivariant representations as well.
This enables the efficient propagation of directional information by scaling linearly with the number of neighbors and keeping the required cutoff small.

\new{Note that neither many-body representations, using angles, dihedral angles etc., nor equivariant representations corresponding to a multipole expansion are complete.
It has been shown that even when including up to 4-body invariants, there are structures with as few as eight atoms which cannot be distinguished \cite{pozdnyakov2020incompleteness}{}  and that a many-body expansion up to $n$-body contributions is necessary to guarantee convergence for a system consisting of $n$ atoms \cite{hermann2007convergence}. The same holds for multipole expansions, where scalars and vectors correspond to the 0th and 1st order.
Thus, even spherical harmonics expansions with less than infinite degree are unable to represent arbitrary equivariant n-body functions.
Instead, a practically sufficient and computationally efficient approach for the problem at hand is desirable.}

In the following, we propose a neural network architecture that takes advantage of these properties. 
It overcomes the limitations of invariant representations discussed above, which we will demonstrate at the example of an organometallic compound in Section~\ref{sec:ferrocene}.

\section{Polarizable atom interaction neural network (\painn{})}
\begin{figure*}[tb]
\centering
\includegraphics[width=0.7\textwidth]{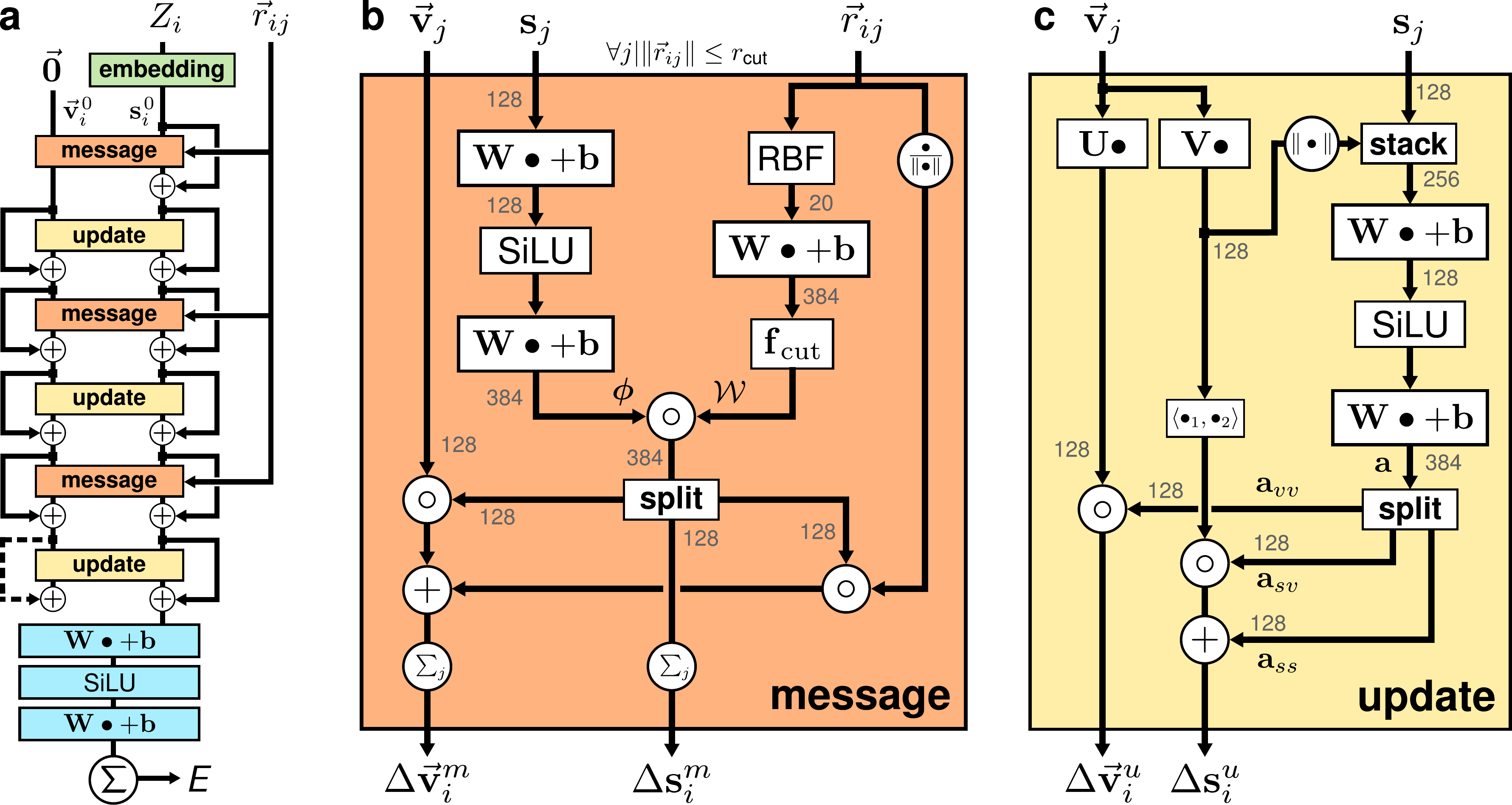}
\caption{\raggedright The architecture of \painn{} with the full architecture (\textbf{a}) as well as the message (\textbf{b}) and update blocks (\textbf{c}) of the equivariant message passing. In all experiments, we use 128 features for $\q_i$ and $\mmu_i$ throughout the architecture. Other layer sizes are annotated in grey.\label{fig:painn}}
\end{figure*}

The potential energy surface $E(Z_1, \dots, Z_N, \rr_1, \dots, \rr_N)$, with nuclear charges $Z_i \in \mathbb{N}$ and atom positions $\rr_i \in \mathbb{R}^3$ exhibits certain symmetries.
Generally, they include invariance of the energy towards the permutation of atom indices $i$, as well as rotations and translations of the molecule.
A neural network potential should encode these constraints to ensure the symmetries of the predicted energy surface and increase data efficiency.
A common inductive bias of the neural network potential is a decomposition of the energy into atomwise contributions 
$E = \sum_{i=1}^{N} \epsilon(\mathbf{s}_i)$,
where an output network $\epsilon$ predicts energy contributions from atoms embedded within their chemical environment, represented by $\mathbf{s}_i \in \mathbb{R}^F$~\cite{behler2007generalized,bartok2013representing}.

While properties of chemical compounds may be such rotationally invariant scalars, they can also be equivariant tensorial properties, e.g. the multipole expansion of the electron density 
\begin{align}
n(\rr) = q + \vec{\mu}^\top \rr + \rr^\top Q \rr + \dots, \label{eq:mpexp}
\end{align}
with charge $q$, dipole $\vec{\mu}$, quadrupole $Q$ and so on.
Similarly, one can interpret invariant and equivariant atomwise representations as local charges and dipole moments of atom-centered multipole expansions~\cite{gastegger2020machine}.
We leverage this in Section~\ref{sec:tensprop} to predict tensorial molecular properties.
We coin our proposed architecture \emph{polarizable atom interaction neural network} (\painn{}).
%Fig.~\ref{fig:painn} provides an overview of the architecture.

\subsection{Representation}
The inputs to \painn{} are the nuclear charges $Z_i \in \mathbb{N}$ and positions $\rr_i \in \mathbf{R}^3$ for each atom $i$.
Similar to previous approaches, the invariant atom representations are initialized to learned embeddings of the atom type $\q_i^0 = \mathbf{a}_{Z_i} \in \mathbb{R}^{F \times 1}$. 
We keep the number of features $F$ constant throughout the network.
The equivariant representations are set to $\mmu_i^0 = \vec{\mathbf{0}} \in \mathbb{R}^{F \times 3}$, since there is no directional information available initially.

Next, we define message and update functions as introduced in Sec.~\ref{sec:eqmpnn}.
We use a residual structure of interchanging message and update blocks (Fig.~\ref{fig:painn}\textbf{a}), resulting in coupled scalar and vectorial representations.
For the residual of the scalar message function, we adopt the feature-wise, continuous-filter convolutions introduced by \citet{schutt2017schnet}
\begin{align}
\Delta \q_i^m &= (\bm{\phi}_{s}(\q) * \mathcal{W}_{s})_i \\ 
&= \sum_{j} \bm{\phi}_{s}(\q_j) \circ \mathcal{W}_{s}(\|\rr_{ij}\|), \notag
\end{align}
where $\phi_s$ consists of atomwise layers as shown in Fig.~\ref{fig:painn}\textbf{b}.
The roationally-invariant filters $\mathcal{W}_s$ are linear combinations of radial basis functions 
$ \sin(\frac{n \pi}{r_\text{cut}} \| \rr_{ij} \|) / \| \rr_{ij} \| $ as proposed by \citet{klicpera2020directional} with $1 \leq n \leq 20$.
Additionally, we apply a cosine cutoff to the filters~\cite{behler2011atom}.

Analogously, we use continuous-filter convolutions for the residual of the equivariant message function
\begin{align}
\Delta \mmu_i^m = &\sum_{j} \mmu_j \circ \bm{\phi}_{vv}(\q_j) \circ \mathcal{W}_{vv}(\|\rr_{ij}\|) \label{eq:msgv} \\
+ &\sum_{j}\bm{\phi}_{vs}(\q_j) \circ \mathcal{W}'_{vs}(\|\rr_{ij}\|) \frac{\rr_{ij}}{\|\rr_{ij}\|}, \notag
\end{align}
where the first term is a convolution of an invariant filter with scaled, equivariant features $\sum_{j} \mmu_j \circ \bm{\phi}_s(\q_j)$, related to the gating proposed by \citet{weiler20183d} as an equivariant nonlinearity.
This propagates directional information obtained in previous message passes to neighboring atoms.
The second term is a convolution of invariant features with an equivariant filter.
This can be derived as the gradient of an invariant filter 
\[
\nabla \mathcal{W}_{vs}(\|\rr_{ij}\|) = \mathcal{W}'_{vs}(\|\rr_{ij}\|) \frac{\rr_{ij}}{\|\rr_{ij}\|}.
\]
Since $\mathcal{W}'_{vs}(\|\rr_{ij}\|)$ is another invariant filter, we can model it directly without taking the derivative.
Furthermore, we use a shared network $\phi$ to perform the transform of all message functions and split the features afterwards (see Fig.~\ref{fig:painn}\textbf{b}).

After the features-wise message blocks, the update blocks are applied atomwise across features.
The residual of the scalar update function is given by
\begin{align}
\Delta \q_i^u &= \mathbf{a}_{ss}(\q_i, \| \mathbf{V} \mmu_i \|) \label{eq:updv} \\ 
&+ \mathbf{a}_{sv}(\q_i, \| \mathbf{V} \mmu_i \|) \, \langle \mathbf{U} \mmu_i, \mathbf{V} \mmu_i \rangle. \notag
\end{align}
Again, we use a scaling functions computed by a shared network $\mathbf{a}(\q_i, \| \mathbf{V} \mmu_i \|)$ as nonlinearity.
In this case, the norm of a linear combination of features is also used to obtain the scaling, thus, coupling the scalar representations with contracted equivariant features.
In a second term, we use the scalar product of two linear combinations of equivariant features.
Similarly, we define the residual for the equivariant features
\begin{align}
\Delta \mmu_i^u &= \mathbf{a}_{vv}(\q_i, \| \mathbf{V} \mmu_i \|) \,  \mathbf{U} \mmu_i,
\end{align}
which is again a nonlinear scaling of linearly combined equivariant features.

\subsection{Prediction of tensorial properties}\label{sec:tensprop}
\begin{figure}[tb]
\centering
\includegraphics[width=0.45\columnwidth]{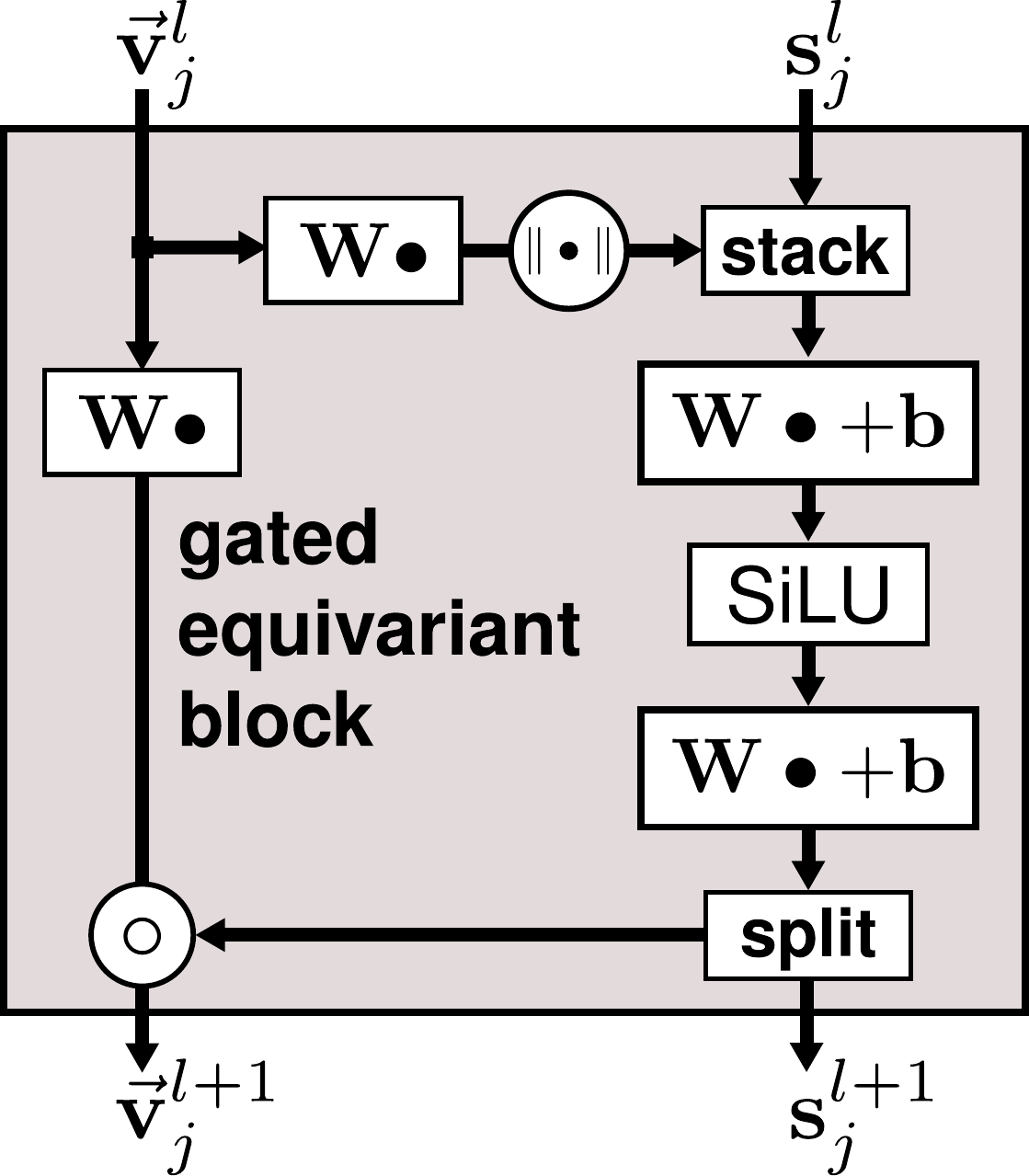}
\caption{Gated equivariant block.\label{fig:geb}}
\end{figure}

\painn{} is able to predict scalar properties using atomwise output layers of rotationally invariant representations $\q_i$, in the same manner as previous approaches~\cite{schutt2017deep,unke2019physnet}.
Beyond that, the additional equivariant representations enable the prediction of equivariant tensors of order $M$ that can be constructed using a rank-1 tensor decomposition
\begin{align}
    T &= \sum_{i=1}^N \sum_{k=1}^R \lambda(\q_i)  \vec{\nu}(\mmu_i)_{k,1} \otimes \dots \otimes \vec{\nu}(\mmu_i)_{k,M}. \label{eq:tpred}
\end{align}
The scalar and vectorial components of this decomposition can be obtained from an output network of \emph{gated equivariant blocks}, as shown in Fig.~\ref{fig:geb}.
\new{These components again make use gated equivariant nonlinearities \cite{weiler2018learning} and show similarities to geometric vector perceptrons (GVP)~\cite{jing2021learning}.
However, while GVP blocks are used as message functions, \painn{} keeps the pairwise message functions comparatively lightweight.
Instead complex transformations are restricted to the atomwise update function and tensor output network.}

Depending on the chemical property of interest, one may want to include additional constraints or replace $\vec{\nu}_{k,m}$ with the atom positions to include the global molecular geometry.
% Depending on the tensorial property of interest, one may need to include non-equivariant components in Eq.~\ref{eq:tpred}, e.g. by replacing some of the $\vec{\nu}(\mmu_i)_{k,m}$ with unit vectors to include the reference frame.
In the following, we demonstrate how this can be employed for two tensorial properties that are important for the simulation of molecular spectra.

The molecular dipole moment $\vec{\mu}$ is the response of the molecular energy to an electric field $\nabla_{\vec{F}}  E$  and, at the same time, the first moment of the electron density (see Eq.~\ref{eq:mpexp}).
It is often predicted using latent atomic charges~\cite{gastegger2017machine}: 
\begin{equation}
\vec{\mu} = \sum_{i=1}^N q_\text{atom}(\q_i) \vec{r_i}. \label{eq:muold}
\end{equation}
This assumes the center of mass at $\rr=\vec{0}$ for brevity.
While this only leverages invariant representations, we can take advantage of the equivariant features of \painn{}.
In this picture, the molecular dipole is constructed from polarized atoms, i.e. both atomic dipoles and charges~\cite{veit2020predicting}, yielding
\begin{align}
    \vec{\mu} = \sum_{i=1}^N \vec{\mu}_\text{atom}(\mmu_i) + q_\text{atom}(\q_i) \vec{r_i}. \label{eq:mu}
\end{align}
In terms of Eq.~\ref{eq:tpred}, this is a simple case with $M=1, R=2$, where the first term predicts the latent local dipoles and the second term computes the dipole generated by the local charges.

As an example with order $M=2$, we consider the polarizability tensor, which describes the response of the molecular dipole to an electric field $J_{\vec{F}}(\vec{\mu}) = H_{\vec{F}}(E)$.
%We construct symmetric atomic polarizability tensors
% \begin{align}
%     \bm{\alpha}_i &= \alpha_\text{0}(\q_i) \, I_3 + \sum_{k=1}^3 \vec{\nu}_r(\mmu_i) \otimes \vec{\nu}(\mmu_i)_r,
% \end{align}
% where the first term is used as a bias for the isotropic component.
% The decomposition of the anisotropic component uses $r=3$ to allow for a matrix with full rank.
% We omit the $lambda$ to constrain the resulting polarizabilities to have positive eigenvalues.
% \citet{wilkins2019accurate} predicted polarizabilities directly as a sum over atomic polarizabilities.
% We include the response of local dipoles, leading to a molecular polarizability tensor
% \begin{align}
%     \bm{\alpha} = &\sum_{i=1}^N \bm{\alpha}_i + \vec{\mu}_\text{atom}(\mmu_i) \odot \vec{r_i}. 
% \end{align}
We construct polarizability tensors using
\begin{align}
    \bm{\alpha} = \sum_{i=1}^N \alpha_\text{0}(\q_i) \, I_3 +\vec{\nu}(\mmu_i) \otimes \vec{r_i} + \vec{r_i} \otimes \vec{\nu}(\mmu_i), \label{eq:alpha}
\end{align}
where the first term of the sum models isotropic, atomwise polarizabilities.
The other two terms add the anisotropic components of the polarizability tensor.
Similar to the charges in Eq.~\ref{eq:mu}, here the atom positions are used to incorporate the global structure of the molecule.

\section{Results}\label{sec:results}
% In the following, we evaluate the performance of \painn{} on a diverse set of organic molecules across chemical compound space, as well as for the accurate prediction of energies and forces of molecular dynamics trajectories on two common benchmarks.
% Then, we study the ability of \painn{} to propagate directional information beyond the chosen cutoff.
% Finally, we apply our model to the simulation of molecular spectra, exploiting its ability to predict tensorial properties.

\painn{} has been implemented using \textsc{PyTorch}~\cite{paszke2019pytorch} and \textsc{SchNetPack}~\cite{schutt2018schnetpack}.
All models use $F=128$ and two output layers as shown in Fig.~\ref{fig:painn} (blue) with the type of output layers depending on the property of interest.
They were trained using the Adam optimizer~\cite{kingma2014adam}, the squared loss and weight decay $\lambda=0.01$~\cite{loshchilov2017decoupled}, if not stated otherwise.
We decay the learning rate by a factor of 0.5 if the validation loss plateaus. We apply exponential smoothing with factor $0.9$ to the validation loss to reduce the impact of fluctuations which are particularly common when training with both energies and forces.
Please refer to the supplement for further details on training parameters.

\subsection{Chemical compound space}
\begin{table*}[tb]
\caption{Mean absolute errors on QM9 dataset for various chemical properties. Results for \painn{} are averaged over three random splits. Best in \textbf{bold}.}
\label{tab:qm9}
\begin{center}
\begin{small}
\begin{tabular}{llrrrrrrr}
\toprule
\textit{Target} & \textit{Unit} & \textsc{\textbf{SchNet}} & \textsc{\textbf{PhysNet}} & \textsc{\textbf{DimeNet++}} & \textsc{\textbf{Cormorant}} & \textsc{\textbf{L1Net}} & \textsc{\textbf{\painn{}}} \\ \midrule
$\mathrm{\mu}$ & D & 0.033 & 0.053 & 0.030 & 0.038 & 0.043 & \textbf{0.012}\\
$\mathrm{\alpha}$ & a$_0^3$ &  0.235  & 0.062 & \textbf{0.044}  & 0.085 & 0.088 & 0.045 \\
$\mathrm{\epsilon}_\text{HOMO}$ & meV & 41 & 32.9 & \textbf{24.6}  & 34 & 46.0 & 27.6 \\
$\mathrm{\epsilon}_\text{LUMO}$ & meV & 34 & 24.7 & \textbf{19.5}  & 38 & 34.6 & 20.4 \\
$\mathrm{\Delta \epsilon}$ & meV & 63 & 42.5 & \textbf{32.6} & 38 & 67.5 &  45.7 \\
$\mathrm{\langle R^2 \rangle}$ & a$_0^2$ & 0.073 & 0.765 & 0.331 & 0.961 & 0.354 & \textbf{0.066} \\
ZPVE & meV & 1.7 & 1.39 & \textbf{1.21} & 2.03 & 1.56 & 1.28 \\
U$_0$ & meV & 14 & 8.15 & 6.32 & 22 & 13.46  & \textbf{5.85} \\
U & meV & 19 & 8.34 & 6.28 & 21 & 13.83  & \textbf{5.83} \\
H & meV & 14 & 8.42 & 6.53 & 21 & 14.36  & \textbf{5.98} \\
G & meV & 14 & 9.40 & 7.56 & 20 & 13.99  & \textbf{7.35} \\
c$_\text{v}$ & $\mathrm{\frac{cal}{mol \, K}}$ & 0.033 & 0.028 & \textbf{0.023} & 0.026 & 0.031 & 0.024 \\
\bottomrule
\end{tabular}
\end{small}
\end{center}
\vskip -0.1in
\end{table*}

We use the QM9 dataset of $\approx$130k small organic molecules~\cite{ramakrishnan2014quantum} with up to nine heavy atoms to evaluate the performance of \painn{} for the prediction of scalar properties across chemical compound space.
We predict the magnitude of the dipole moment using Eq.~\ref{eq:mu} and the electronic spatial extent by
\[
\left\langle R^2 \right\rangle =  \sum_{i=1}^N q_\text{atom}(\q_i) \| \vec{r}_i\|^2,
\]
as implemented by \textsc{SchNetPack}.
The remaining properties are predicted as sums over atomic contributions.
\painn{} is trained on 110k examples while 10k molecules are used as a validation set for decaying the learning rate and early stopping. 
The remaining data is used as test set and results are averaged over three random splits.
For the isotropic polarizability $\alpha$, we first observed validation MAEs of $0.054$~$a_0$.
Upon closer inspection, we notice that for this property both the squared loss as well as the MAE can be reduced when minimizing the MAE directly (as done by \citet{klicpera2020fast}).
This yields both validation and test MAEs of $0.045 a_0$ that are comparable to those of \textsc{DimeNet++}.

Tab.~\ref{tab:qm9} shows the mean absolute error (MAE) of \painn{} for 12 target properties of QM9 in comparison with previous approaches.
\textsc{SchNet}~\cite{schutt2017schnet} and \textsc{PhysNet}~\cite{unke2019physnet} are MPNNs with distance-based interactions, \textsc{DimeNet}++~\cite{klicpera2020fast} includes additional angular information and is an improved variant of \textsc{DimeNet}~\cite{klicpera2020directional}. \textsc{L1Net}~\cite{miller2020relevance} and \textsc{Cormorant}~\cite{anderson2019cormorant} are equivariant neural networks based on spherical harmonics and Clebsch-Gordon coefficients.

\painn{} achieves state-of-the-art results in six target properties and yields comparable results to \textsc{DimeNet}++ on another two targets.
On the remaining properties, \painn{} achieves the second best results after \textsc{DimeNet}++.
Note that \painn{} using about 600k parameters is significantly smaller than \textsc{DimeNet}++ with about 1.8M parameters.
For random batches of 50 molecules from QM9, the inference time is reduced from 45 ms to 13 ms, i.e. an improvement of more than 70\%, when comparing \painn{} to the reference implementation of \textsc{DimeNet}++\footnote{\url{https://github.com/klicperajo/dimenet}} using an NVIDIA V100.

\subsection{Molecular dynamics trajectories}

\begin{table*}[tb]
\caption{Mean absolute errors on MD17 dataset for energy and force predictions in kcal/mol and kcal/mol/{\AA}, respectively. \citet{batzner2021se} only reported force errors for \textsc{NequIP}. Results for \painn{} are averaged over three random splits. Best in \textbf{bold}.}
\label{tab:md17}
\begin{center}
\begin{scriptsize}
\begin{sc}
\begin{tabular}{llrrr|rrrrr}
\toprule
& & \textbf{sGDML} & \textbf{NequIP} & \textbf{\painn{}} & \textbf{SchNet} & \textbf{PhysNet} & \textbf{DimeNet} & \textbf{FCHL19} & \textbf{\painn{}} \\ 
& &  \multicolumn{3}{c}{\textit{trained on forces only}} & \multicolumn{5}{c}{\textit{trained on energies \& forces}} \\
\midrule

\multirow{2}{*}{\textbf{Aspirin}} & \textit{energy} & 0.19 & -- & \textbf{0.167} &  0.37 & 0.230 & 0.204 & 0.182 & \textbf{0.159} \\
 & \textit{forces} & 0.68 & 0.348 & \textbf{0.338} & 1.35 & 0.605 & 0.499 & 0.478 & \textbf{0.371} \\ \midrule
 
 \multirow{2}{*}{\textbf{Ethanol}} & \textit{energy} & 0.07 & -- & \textbf{0.064} & 0.08  & 0.059 & 0.064 & \textbf{0.054} & 0.063 \\
 & \textit{forces} & 0.33 & \textbf{0.208} & 0.224 & 0.39 & 0.160 & 0.230 & \textbf{0.136} & 0.230 \\ 
 \midrule
 
 \multirow{2}{*}{\textbf{Malondialdehyde}} & \textit{energy} & \textbf{0.10} & -- & \textbf{0.100}  &  0.13  & 0.094 & 0.104 & \textbf{0.081} & 0.091 \\
 & \textit{forces} & 0.41 & \textbf{0.337} & 0.344 & 0.66 & 0.319 & 0.383 & \textbf{0.245} & 0.319 \\ 
 \midrule
 
 \multirow{2}{*}{\textbf{Naphthalene}} & \textit{energy} & \textbf{0.12} & -- & \textbf{0.116} & 0.16 & 0.142 & 0.122 & \textbf{0.117} & \textbf{0.117} \\
 & \textit{forces} & 0.11 & 0.096 & \textbf{0.077} & 0.58 & 0.310 & 0.215 &  0.151 & \textbf{0.083} \\ 
 \midrule
 
 \multirow{2}{*}{\textbf{Salicylic Acid}} & \textit{energy} & \textbf{0.12} & -- & \textbf{0.116} & 0.20 & 0.126 & 0.134 & \textbf{0.114} & \textbf{0.114} \\
 & \textit{forces} & 0.28 & 0.238 & \textbf{0.195} & 0.85 & 0.337 & 0.374 & 0.221 &  \textbf{0.209} \\ 
 \midrule
 
 \multirow{2}{*}{\textbf{Toluene}} & \textit{energy} & \textbf{0.10} & -- & \textbf{0.095} & 0.12 & 0.100 & 0.102 & 0.098 & \textbf{0.097} \\
 & \textit{forces} & 0.14 & 0.101 & \textbf{0.094} & 0.57 & 0.191 & 0.216 & 0.203 & \textbf{0.102} \\ 
 \midrule
 
 \multirow{2}{*}{\textbf{Uracil}} & \textit{energy} & \textbf{0.11} & -- & \textbf{0.106} & 0.14 & 0.108 & 0.115 & \textbf{0.104} & \textbf{0.104} \\
 & \textit{forces} & 0.24 & 0.172 & \textbf{0.139} & 0.56 & 0.218 & 0.301 & \textbf{0.105} & 0.140 \\
\bottomrule
\end{tabular}
\end{sc}
\end{scriptsize}
\end{center}
\vskip -0.1in
\end{table*}

We evaluate the ability to predict combined energies and forces on the MD17 benchmark~\cite{chmiela2017machine} including molecular dynamics trajectories of small organic molecules.
While the atomic forces could be predicted directly from vectorial features, we employ the gradients of the energy model $\vec{F}_i = -\partial E / \partial \rr_i $ to ensure conservation of energy.
This property is crucial to run stable molecular dynamics simulations.
To demonstrate the data efficiency of \painn{}, we use the more challenging setting with 1k known structures of which we use 950 for training and 50 for validation, where a separate model is trained for each trajectory.
Tab.~\ref{tab:md17} shows the comparison with \textsc{sGDML}~\cite{chmiela2018} and  \textsc{NequIP}~\cite{batzner2021se}, which were trained on forces only, as well as \textsc{SchNet}, \textsc{PhysNet}, \textsc{DimeNet} and \textsc{FCHL19}~\cite{christensen2020fchl}, that were trained on a combined loss of energies and forces.
\citet{christensen2020role} have found that the energies of MD17 are noisy. Thus, depending on the molecule and chosen tradeoff, using energies for training is not always beneficial.
For this reason, we train two \painn{} models per trajectory: only on forces and on a combined loss including energy with a force error weight of $\rho=0.95$.
\painn{} achieves the lowest mean absolute errors for 12 out 14 targets on models trained only on forces and exhibits errors in a similar range as Gaussian regression with the \textsc{FCHL19} kernel.
Overall, \painn{} performs best or equal to \textsc{FCHL19} on 9 out of 14 targets.
This demonstrates that equivariant neural networks approaches such as \painn{} are able to compete with kernel methods in the small data regime, while being able to scale to large data sets at the same time.

\subsection{Advantages of equivariant features}
\subsubsection{Ablation studies}
\begin{table}
\begin{center}
\begin{scriptsize}
\caption{Ablation study for the prediction of energies [kcal/mol] and forces [kcal/mol/{\AA}] for aspirin trajectories from MD17.\label{tab:abl}}
\begin{tabular}{lllrr} \toprule
\textbf{Ablation}    & \textbf{\textit{\# params}} & \textbf{$F$} & \textbf{\textit{energy MAE}} & \textbf{\textit{force MAE}} \\ \midrule
no ablation & 588.3k & 128 & 0.159 & 0.371 \\ \midrule
no scalar product of & 589.1k & 134 & 0.173 & 0.420 \\
vector features in Eq. 8 \\
no vector propagation   & 589.2k & 135 & 0.183 & 0.441 \\
($\mathcal{W}_{vv} = 0$ in Eq. 7) \\
remove both & 590.1k & 142  & 0.200 & 0.507 \\  \midrule
no vector features & 590.3k & 174 & 0.449 & 1.194 \\ 
\bottomrule
\end{tabular}
\end{scriptsize}
\end{center}
\end{table}
\new{We evaluate the impact of equivariant vector features at the example of the aspirin MD trajectory from the previous section.
Compared to the full model, we remove the scalar product of vector features in Eq.~\ref{eq:updv} from the update block and the convolution over vector features in Eq.~\ref{eq:msgv} (i.e., $\mathcal{W}_{vv} = 0$).
Table~\ref{tab:abl} show the results for the various ablations.
The number of parameters is kept approximately constant by raising the number of node features $F$ accordingly.
We observe that all ablated components contribute to the final accuracy of the model, where the convolution over equivariant features in the message function has a slightly larger impact.
This component also enables the propagation of directional information, which will be examined in Sec.~\ref{sec:ferrocene}.
Finally, we remove all vector features from the model, resulting in an invariant model.
Despite keeping the number of parameters constant by increasing the number of atoms features to $F=174$, the mean absolute error of the forces increases beyond 1 kcal/mol/{\AA}.
}

\subsubsection{Propagation of directional information in substituted ferrocene}\label{sec:ferrocene}
\begin{figure}[tb]
\centering
\includegraphics[width=.97\columnwidth]{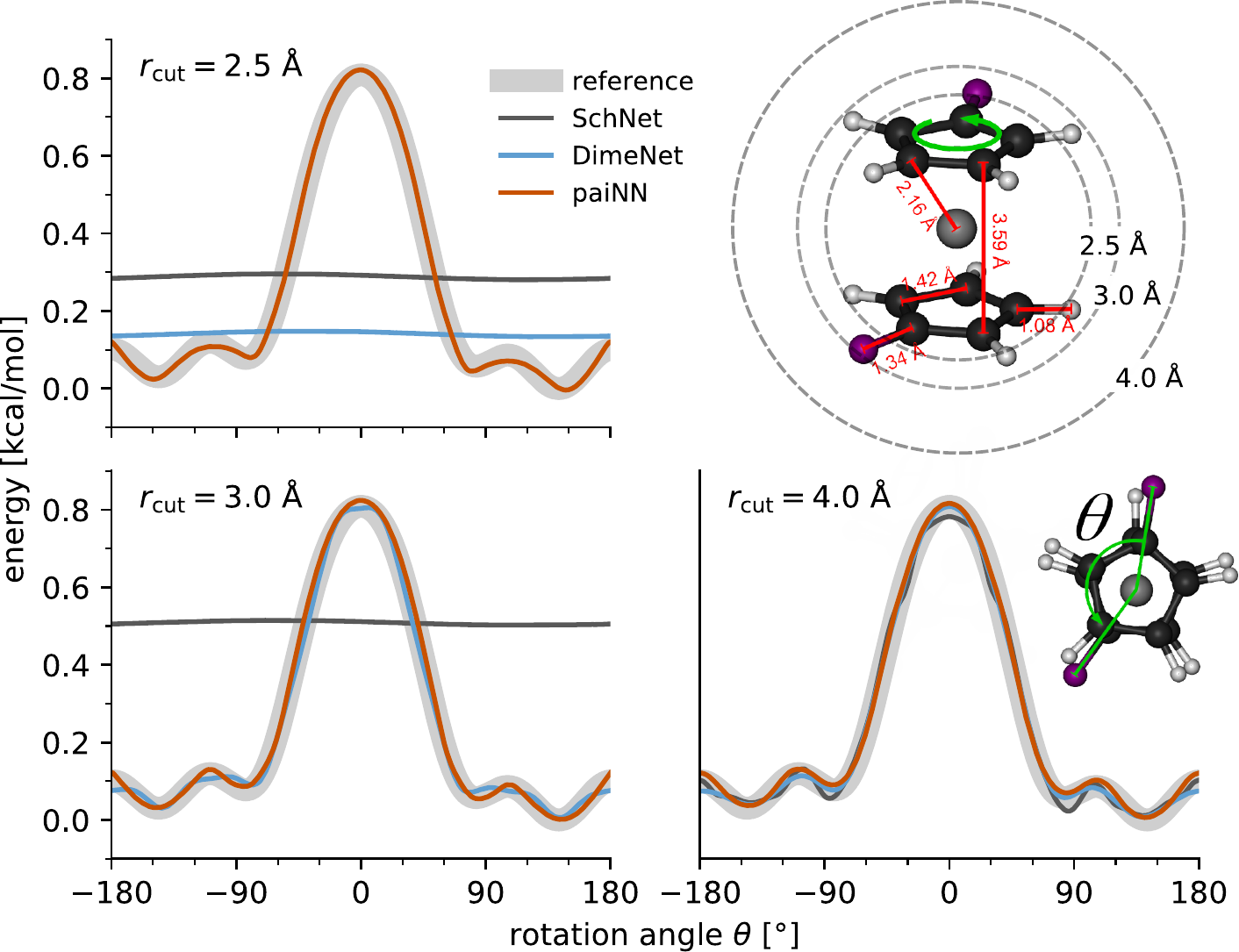}
\caption{Rotational energy profile of substituted ferrocene obtained by varying the rotation angle~$\theta$ while keeping all bond distances fixed at their equilibrium values (H: white, C: black, F: purple, Fe: grey). For small cutoff $r_{\rm cut}$ (dashed grey circles), MPNNs with scalar feature representations are unable to represent information about the rotation angle~$\theta$.}
%and consequently predict a flat energy profile.}
\label{fig:ferrocene_rotation}
\end{figure}

To demonstrate the advantages of equivariant over invariant representations in practice, we consider a ferrocene derivative where one hydrogen atom in each cyclopentadienyl ring is substituted by fluorine (see Fig.~\ref{fig:ferrocene_rotation}).
This molecule has been chosen as it features small energy fluctuations ($<$1~kcal/mol) when the rings rotate relative to each other. 
Since the torsional energy profile depends mainly on the orientation of the distant fluorine atoms (measured by the rotation angle $\theta$), it is a challenging prediction target for models without equivariant representations.
Fig.~\ref{fig:ferrocene_rotation} shows the predicted energy profiles for a full rotation of the cyclopentadienyl ring using cutoffs $r_{\rm cut} \in \{ 2.5, 3.0, 4.0 \}$~\AA.
All models are trained on energies of 10k structures sampling thermal fluctuations (300~K) and ring rotations of substituted ferrocene.
Another 1k structures are used for validation.
SchNet~\cite{schutt2017schnet}, which uses scalar features, predicts a flat energy profile for $r_{\rm cut} \leq 3.0$~\AA, because it is unable to resolve $\theta$.
Although it is possible to encode the relevant information when using a larger cutoff, the learned representation seems to be susceptible to noise, as indicated by the deviations of the predicted energy profile. 
Since DimeNet~\cite{klicpera2020directional} includes terms that explicitly depend on angles between triplets of atoms, $\theta$ can be resolved for smaller cutoffs, but still fails for $r_{\rm cut} = 2.5$~\AA.
In contrast, the equivariant environment representation of \painn{} allows to faithfully reproduce the torsional energy profile even for very small cutoffs.

\subsection{Molecular Spectra}

\begin{figure*}[tb]
    \centering
    \includegraphics[width=0.9\textwidth]{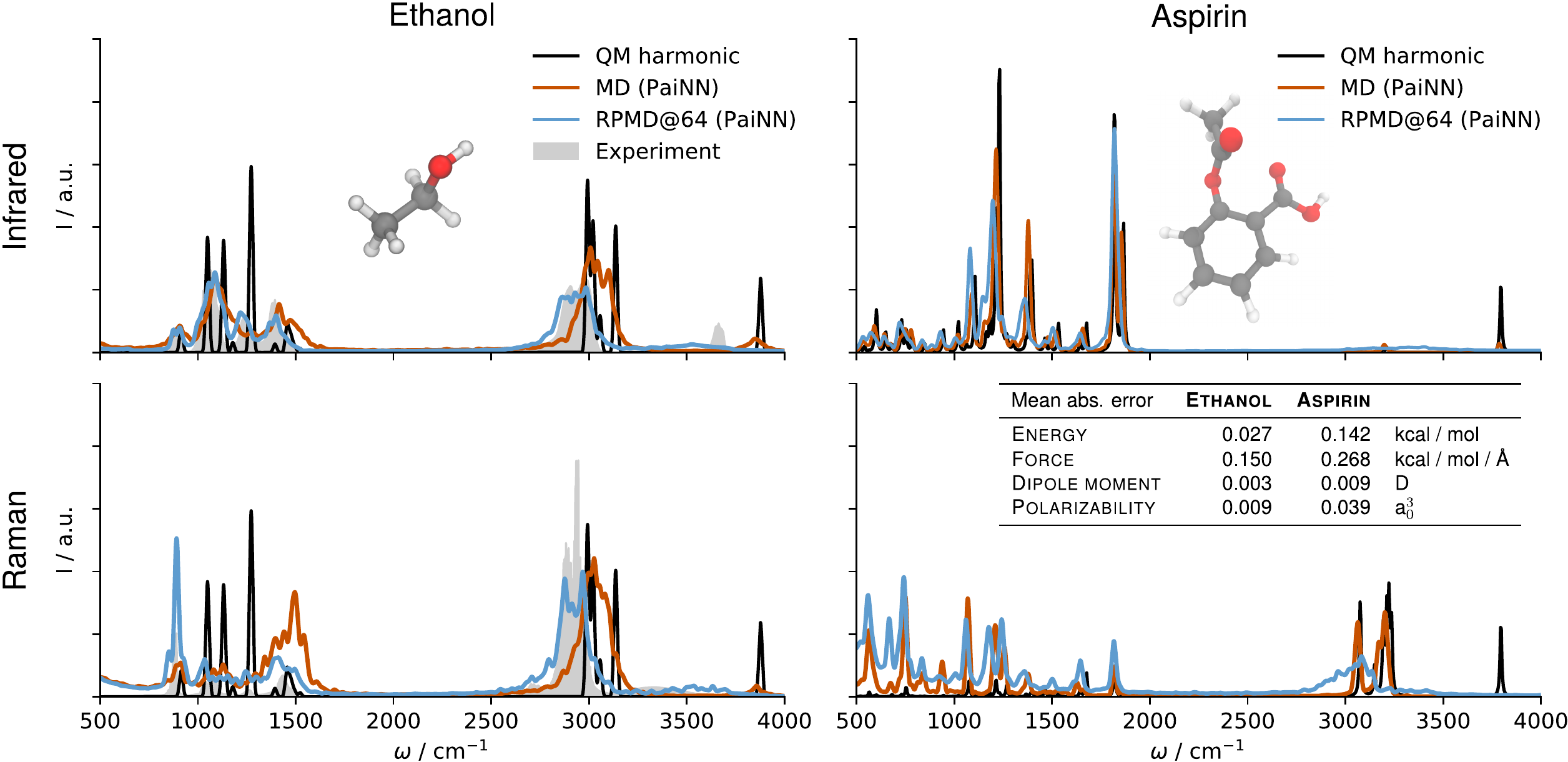}
    \caption{IR (top) and Raman (bottom) spectra of ethanol and aspirin. \new{Spectra calculated with the reference method using the harmonic oscillator approximation are shown in black (QM harmonic).} The inset table shows the mean absolute errors on the respective test set.}
    \label{fig:spectra_aspirin}
\end{figure*}

We apply \painn{} to the efficient computation of infrared and Raman spectra of ethanol and aspirin.
%from molecular dynamics (MD) simulations
%The equivariant \painn{} representations enable the direct prediction of vectorial and tensorial properties of molecules and materials.
%When applied to molecular dipole moments and polarizability tensors, \painn{} can be used to efficiently obtain infrared and Raman spectra from molecular dynamics (MD) simulations.
\new{Although these spectra can in principle be obtained from a single molecular structure via the harmonic oscillator approximation, such an approach tends to neglect important effects, e.g. different molecular conformations~\cite{thomas2013computing}.
In order to obtain high quality spectra,}
molecular dynamics simulations have to be performed, requiring the prediction of forces at each time step.
Additionally, one needs to compute dipole moments and polarizabilities along the generated trajectory.
The infrared and Raman spectra are obtained as the Fourier transform of the time auto-correlation functions of the respective property.
Using electronic structure methods, such simulations become prohibitively expensive due to the large number of successive computations required.
Since nuclear quantum effects (NQE) need to be considered to obtain high quality spectra~\cite{sauceda2021dynamical}, ring-polymer molecular dynamics (RPMD) simulations need to be performed, which treat multiple copies of a molecule simultaneously and further increase computational cost.
 
%Here, we demonstrate how these simulations can be accelerated with \painn{}, using it to predict the infrared and Raman spectra . 
We train a joined model for energies, forces, dipole moments and polarizability tensors on 8k ethanol conformations using additional 1k molecules each as validation and test set.
A second model is trained for aspirin with 16k training examples and 2k molecules for validation and testing.
Energies and forces are predicted as described above for MD17.
The dipole moments and polarizability tensors are obtained as described in Section~\ref{sec:tensprop}, employing two gated equivariant blocks each yielding atomwise scalars and vectors, that are used to compute the outputs as described in Eqs.~\ref{eq:mu} and \ref{eq:alpha}.
The joined model exhibits accurate predictions for energy, forces, dipole moments and polarizabilities, as shown in the inset table of Fig~\ref{fig:spectra_aspirin}.
\new{To evaluate the importance of equivariant features for the dipole prediction, we have trained an alternative model using only latent partial charges, but no local dipoles. The results show that predicting molecular dipoles using vector features (Eq.~\ref{eq:mu}) reduces the mean abs. error by more than 50\%, reaching 0.009~D compared to 0.020~D using only scalars and positions (Eq.~\ref{eq:muold}).}
Simulated infrared and Raman spectra using classical MD and RPMD with 64 replicas \new{and the respective \painn{} models} are shown for both molecules in Fig.~\ref{fig:spectra_aspirin}, alongside \new{harmonic} spectra obtained with the electronic structure reference and, for ethanol, experimental spectra recorded in the gas phase~\cite{NIST,kiefer2017simultaneous}.

The peak positions and intensities of ethanol infrared spectra (Fig.~\ref{fig:spectra_aspirin}a, top) computed with classical MD and \new{\painn{}} agree closely with the static electronic structure reference, e.g. in the C-H and O-H stretching regions at 3000~\icm{} and 3900~\icm{}.
This indicates, that \painn{} faithfully reproduces the original electronic structure method in general.
However, when compared to experiment, both spectra are shifted towards higher frequencies.
The \painn{} RPMD spectrum, on the other hand shows excellent agreement with experiment, illustrating the importance of NQEs.
Similar observations can be made for the ethanol Raman spectrum (Fig.~\ref{fig:spectra_aspirin}a, bottom), where the \painn{} RMPD spectrum once again offers the most faithful reproduction of the experiment.
% aspirin
We observe similar trends for aspirin (Fig.~\ref{fig:spectra_aspirin}b).
RPMD spectra are once again shifted to lower frequencies while the classical MD closely resembles the static electronic structure spectrum.
% the following can be cut if necessary
% The aspirin spectra further illustrate that access to both infrared and Raman spectra is advantageous, as they offer complementary information.
% Several bands in the bond stretching region close to 3000~\icm{} are absent in the infared spectrum (top) but active in the Raman spectrum (bottom) and \emph{vice versa}, although both were obtained from the same simulation.

\painn{} reduces the time required to obtain these spectra by several orders of magnitude.
For ethanol, an individual reference computation takes $\approx$ 140~seconds compared to 14~ms for a \painn{} evaluation on a V100 GPU.
This corresponds to a reduction of the overall simulation time from 400 days to approximately one hour.
The speedup offered by \painn{} is even more pronounced for aspirin, where a simulation that would have taken 25 years (3140~seconds / step) now takes one hour (15~ms / step).
%Beyond that \painn{} leverages the parallelism of GPUs to carry out RPMD simulations using 64 replicas in the same time that it requires for the conventional MD simulations.
% We train a joined model for the prediction of energies, forces, dipole moments and polarizability tensors on 16k aspirin conformations using additional 2k molecules each as validation and test set.
% The energies and forces are predicted as described above for MD17.
% The dipole moments and polarizability tensors are obtained as described in Section~\ref{sec:tensprop}, employing two gated equivariant blocks each yielding atomwise scalars and vectors, that are used to compute the outputs as described in Eqs.~\ref{eq:mu} and \ref{eq:alpha}.
% The joined model exhibits accurate predictions with MAEs of \note{xxx}, for energy, forces, dipole moments and polarizabilities, respectively.

\section{Conclusions}
We have given general guidelines to design equivariant MPNNs and discussed the advantages of equivariant representations over angular features in terms of computational efficiency as well as their ability to propagate directional information.
On this basis, we have proposed \painn{} that yields fast and accurate predictions of scalar and tensorial molecular properties.
Thereby, equivariant message passing allows us to significantly reduce both model size and inference time compared to directional message-passing while retaining accuracy.
Finally, we have demonstrated that \painn{} can be applied to the prediction of tensorial properties, which we leverage to accelerate the simulation of molecular spectra by 4-5 orders of magnitude -- from years to hours.

In future work, the equivariant representation of \painn{} as well as the ability to predict tensorial properties may be leveraged in generative models of 3d geometries~\cite{gebauer2019symmetry,kohler2019equivariant,simm2020symmetry} or the prediction of wavefunctions~\cite{hegde2017machine,schutt2019unifying,hermann2020deep}.
We see further applications of equivariant message passing in 3d shape recognition and graph embedding~\cite{goyal2018graph}.

Many challenges remain for the fast and accurate prediction of molecular properties, e.g.  modeling of enzymatic active sites or surface reactions.
To describe such phenomena, highly accurate reference methods are required. Due to their computational cost, reference data generation can become a bottleneck, making data-efficient MPNNs such as \painn{} invaluable for future chemistry research.

\section*{Acknowledgements}
KTS acknowledges support by the Federal Ministry of Education and Research (BMBF) for the Berlin Center for Machine Learning / BIFOLD (01IS18037A).
MG works at the BASLEARN – TU Berlin/BASF Joint Lab for Machine Learning, co-financed by TU Berlin and BASF SE.
OTU acknowledges funding from the Swiss National Science Foundation (Grant No. P2BSP2\_188147).

\bibliography{references}

%aipnum4-2.bst 2019-01-14 (MD) hand-edited version of apsrev4-1.bst
\begin{thebibliography}{69}%
\makeatletter
\providecommand \@ifxundefined [1]{%
 \@ifx{#1\undefined}
}%
\providecommand \@ifnum [1]{%
 \ifnum #1\expandafter \@firstoftwo
 \else \expandafter \@secondoftwo
 \fi
}%
\providecommand \@ifx [1]{%
 \ifx #1\expandafter \@firstoftwo
 \else \expandafter \@secondoftwo
 \fi
}%
\providecommand \natexlab [1]{#1}%
\providecommand \enquote  [1]{``#1''}%
\providecommand \bibnamefont  [1]{#1}%
\providecommand \bibfnamefont [1]{#1}%
\providecommand \citenamefont [1]{#1}%
\providecommand \href@noop [0]{\@secondoftwo}%
\providecommand \href [0]{\begingroup \@sanitize@url \@href}%
\providecommand \@href[1]{\@@startlink{#1}\@@href}%
\providecommand \@@href[1]{\endgroup#1\@@endlink}%
\providecommand \@sanitize@url [0]{\catcode `\\12\catcode `\$12\catcode
  `\&12\catcode `\#12\catcode `\^12\catcode `\_12\catcode `\%12\relax}%
\providecommand \@@startlink[1]{}%
\providecommand \@@endlink[0]{}%
\providecommand \url  [0]{\begingroup\@sanitize@url \@url }%
\providecommand \@url [1]{\endgroup\@href {#1}{\urlprefix }}%
\providecommand \urlprefix  [0]{URL }%
\providecommand \Eprint [0]{\href }%
\providecommand \doibase [0]{https://doi.org/}%
\providecommand \selectlanguage [0]{\@gobble}%
\providecommand \bibinfo  [0]{\@secondoftwo}%
\providecommand \bibfield  [0]{\@secondoftwo}%
\providecommand \translation [1]{[#1]}%
\providecommand \BibitemOpen [0]{}%
\providecommand \bibitemStop [0]{}%
\providecommand \bibitemNoStop [0]{.\EOS\space}%
\providecommand \EOS [0]{\spacefactor3000\relax}%
\providecommand \BibitemShut  [1]{\csname bibitem#1\endcsname}%
\let\auto@bib@innerbib\@empty
%</preamble>
\bibitem [{\citenamefont {Behler}(2016)}]{behler2016perspective}%
  \BibitemOpen
  \bibfield  {author} {\bibinfo {author} {\bibfnamefont {J.}~\bibnamefont
  {Behler}},\ }\bibfield  {title} {\enquote {\bibinfo {title} {{P}erspective:
  {M}achine learning potentials for atomistic simulations},}\ }\href@noop {}
  {\bibfield  {journal} {\bibinfo  {journal} {J. Chem. Phys}\ }\textbf
  {\bibinfo {volume} {145}},\ \bibinfo {pages} {170901} (\bibinfo {year}
  {2016})}\BibitemShut {NoStop}%
\bibitem [{\citenamefont {Unke}\ \emph {et~al.}(2020)\citenamefont {Unke},
  \citenamefont {Chmiela}, \citenamefont {Sauceda}, \citenamefont {Gastegger},
  \citenamefont {Poltavsky}, \citenamefont {Sch{\"u}tt}, \citenamefont
  {Tkatchenko},\ and\ \citenamefont {M{\"u}ller}}]{Unke2020mlforce}%
  \BibitemOpen
  \bibfield  {author} {\bibinfo {author} {\bibfnamefont {O.~T.}\ \bibnamefont
  {Unke}}, \bibinfo {author} {\bibfnamefont {S.}~\bibnamefont {Chmiela}},
  \bibinfo {author} {\bibfnamefont {H.~E.}\ \bibnamefont {Sauceda}}, \bibinfo
  {author} {\bibfnamefont {M.}~\bibnamefont {Gastegger}}, \bibinfo {author}
  {\bibfnamefont {I.}~\bibnamefont {Poltavsky}}, \bibinfo {author}
  {\bibfnamefont {K.~T.}\ \bibnamefont {Sch{\"u}tt}}, \bibinfo {author}
  {\bibfnamefont {A.}~\bibnamefont {Tkatchenko}},\ and\ \bibinfo {author}
  {\bibfnamefont {K.-R.}\ \bibnamefont {M{\"u}ller}},\ }\bibfield  {title}
  {\enquote {\bibinfo {title} {{M}achine {L}earning {F}orce {F}ields},}\
  }\href@noop {} {\bibfield  {journal} {\bibinfo  {journal} {arXiv preprint
  arXiv:2010.07067}\ } (\bibinfo {year} {2020})}\BibitemShut {NoStop}%
\bibitem [{\citenamefont {von Lilienfeld}, \citenamefont {M{\"u}ller},\ and\
  \citenamefont {Tkatchenko}(2020)}]{von2020exploring}%
  \BibitemOpen
  \bibfield  {author} {\bibinfo {author} {\bibfnamefont {O.~A.}\ \bibnamefont
  {von Lilienfeld}}, \bibinfo {author} {\bibfnamefont {K.-R.}\ \bibnamefont
  {M{\"u}ller}},\ and\ \bibinfo {author} {\bibfnamefont {A.}~\bibnamefont
  {Tkatchenko}},\ }\bibfield  {title} {\enquote {\bibinfo {title} {Exploring
  chemical compound space with quantum-based machine learning},}\ }\href@noop
  {} {\bibfield  {journal} {\bibinfo  {journal} {Nat. Rev. Chem.}\ }\textbf
  {\bibinfo {volume} {4}},\ \bibinfo {pages} {347--358} (\bibinfo {year}
  {2020})}\BibitemShut {NoStop}%
\bibitem [{\citenamefont {Chmiela}\ \emph {et~al.}(2018)\citenamefont
  {Chmiela}, \citenamefont {Sauceda}, \citenamefont {M{\"u}ller},\ and\
  \citenamefont {Tkatchenko}}]{chmiela2018}%
  \BibitemOpen
  \bibfield  {author} {\bibinfo {author} {\bibfnamefont {S.}~\bibnamefont
  {Chmiela}}, \bibinfo {author} {\bibfnamefont {H.~E.}\ \bibnamefont
  {Sauceda}}, \bibinfo {author} {\bibfnamefont {K.-R.}\ \bibnamefont
  {M{\"u}ller}},\ and\ \bibinfo {author} {\bibfnamefont {A.}~\bibnamefont
  {Tkatchenko}},\ }\bibfield  {title} {\enquote {\bibinfo {title} {Towards
  exact molecular dynamics simulations with machine-learned force fields},}\
  }\href {https://doi.org/10.1038/s41467-018-06169-2} {\bibfield  {journal}
  {\bibinfo  {journal} {Nat. Commun.}\ }\textbf {\bibinfo {volume} {9}},\
  \bibinfo {pages} {3887} (\bibinfo {year} {2018})}\BibitemShut {NoStop}%
\bibitem [{\citenamefont {Westermayr}, \citenamefont {Gastegger},\ and\
  \citenamefont {Marquetand}(2020)}]{westermayr2020combining}%
  \BibitemOpen
  \bibfield  {author} {\bibinfo {author} {\bibfnamefont {J.}~\bibnamefont
  {Westermayr}}, \bibinfo {author} {\bibfnamefont {M.}~\bibnamefont
  {Gastegger}},\ and\ \bibinfo {author} {\bibfnamefont {P.}~\bibnamefont
  {Marquetand}},\ }\bibfield  {title} {\enquote {\bibinfo {title} {{C}ombining
  {SchNet} and {SHARC}: {T}he {SchNarc} machine learning approach for
  excited-state dynamics},}\ }\href@noop {} {\bibfield  {journal} {\bibinfo
  {journal} {J. Phys. Chem. Lett.}\ }\textbf {\bibinfo {volume} {11}},\
  \bibinfo {pages} {3828--3834} (\bibinfo {year} {2020})}\BibitemShut {NoStop}%
\bibitem [{\citenamefont {Morawietz}\ \emph {et~al.}(2016)\citenamefont
  {Morawietz}, \citenamefont {Singraber}, \citenamefont {Dellago},\ and\
  \citenamefont {Behler}}]{morawietz2016van}%
  \BibitemOpen
  \bibfield  {author} {\bibinfo {author} {\bibfnamefont {T.}~\bibnamefont
  {Morawietz}}, \bibinfo {author} {\bibfnamefont {A.}~\bibnamefont
  {Singraber}}, \bibinfo {author} {\bibfnamefont {C.}~\bibnamefont {Dellago}},\
  and\ \bibinfo {author} {\bibfnamefont {J.}~\bibnamefont {Behler}},\
  }\bibfield  {title} {\enquote {\bibinfo {title} {{H}ow van der {W}aals
  interactions determine the unique properties of water},}\ }\href@noop {}
  {\bibfield  {journal} {\bibinfo  {journal} {Proc. Natl. Acad. Sci.}\ }\textbf
  {\bibinfo {volume} {113}},\ \bibinfo {pages} {8368--8373} (\bibinfo {year}
  {2016})}\BibitemShut {NoStop}%
\bibitem [{\citenamefont {Bart{\'o}k}\ \emph {et~al.}(2018)\citenamefont
  {Bart{\'o}k}, \citenamefont {Kermode}, \citenamefont {Bernstein},\ and\
  \citenamefont {Cs{\'a}nyi}}]{bartok2018machine}%
  \BibitemOpen
  \bibfield  {author} {\bibinfo {author} {\bibfnamefont {A.~P.}\ \bibnamefont
  {Bart{\'o}k}}, \bibinfo {author} {\bibfnamefont {J.}~\bibnamefont {Kermode}},
  \bibinfo {author} {\bibfnamefont {N.}~\bibnamefont {Bernstein}},\ and\
  \bibinfo {author} {\bibfnamefont {G.}~\bibnamefont {Cs{\'a}nyi}},\ }\bibfield
   {title} {\enquote {\bibinfo {title} {Machine learning a general-purpose
  interatomic potential for silicon},}\ }\href@noop {} {\bibfield  {journal}
  {\bibinfo  {journal} {Phys. Rev. X}\ }\textbf {\bibinfo {volume} {8}},\
  \bibinfo {pages} {041048} (\bibinfo {year} {2018})}\BibitemShut {NoStop}%
\bibitem [{\citenamefont {Lu}\ \emph {et~al.}(2020)\citenamefont {Lu},
  \citenamefont {Wang}, \citenamefont {Chen}, \citenamefont {Liu},
  \citenamefont {Lin}, \citenamefont {Car}, \citenamefont {Jia}, \citenamefont
  {Zhang} \emph {et~al.}}]{lu202086}%
  \BibitemOpen
  \bibfield  {author} {\bibinfo {author} {\bibfnamefont {D.}~\bibnamefont
  {Lu}}, \bibinfo {author} {\bibfnamefont {H.}~\bibnamefont {Wang}}, \bibinfo
  {author} {\bibfnamefont {M.}~\bibnamefont {Chen}}, \bibinfo {author}
  {\bibfnamefont {J.}~\bibnamefont {Liu}}, \bibinfo {author} {\bibfnamefont
  {L.}~\bibnamefont {Lin}}, \bibinfo {author} {\bibfnamefont {R.}~\bibnamefont
  {Car}}, \bibinfo {author} {\bibfnamefont {W.}~\bibnamefont {Jia}}, \bibinfo
  {author} {\bibfnamefont {L.}~\bibnamefont {Zhang}}, \emph {et~al.},\
  }\bibfield  {title} {\enquote {\bibinfo {title} {86 {PFLOPS} {D}eep
  {P}otential {M}olecular {D}ynamics simulation of 100 million atoms with ab
  initio accuracy},}\ }\href@noop {} {\bibfield  {journal} {\bibinfo  {journal}
  {arXiv preprint arXiv:2004.11658}\ } (\bibinfo {year} {2020})}\BibitemShut
  {NoStop}%
\bibitem [{\citenamefont {Gilmer}\ \emph {et~al.}(2017)\citenamefont {Gilmer},
  \citenamefont {Schoenholz}, \citenamefont {Riley}, \citenamefont {Vinyals},\
  and\ \citenamefont {Dahl}}]{gilmer2017neural}%
  \BibitemOpen
  \bibfield  {author} {\bibinfo {author} {\bibfnamefont {J.}~\bibnamefont
  {Gilmer}}, \bibinfo {author} {\bibfnamefont {S.~S.}\ \bibnamefont
  {Schoenholz}}, \bibinfo {author} {\bibfnamefont {P.~F.}\ \bibnamefont
  {Riley}}, \bibinfo {author} {\bibfnamefont {O.}~\bibnamefont {Vinyals}},\
  and\ \bibinfo {author} {\bibfnamefont {G.~E.}\ \bibnamefont {Dahl}},\
  }\bibfield  {title} {\enquote {\bibinfo {title} {Neural message passing for
  quantum chemistry},}\ }\href@noop {} {\bibfield  {journal} {\bibinfo
  {journal} {Interational Conference on Machine Learning}\ } (\bibinfo {year}
  {2017})}\BibitemShut {NoStop}%
\bibitem [{\citenamefont {Chmiela}\ \emph {et~al.}(2017)\citenamefont
  {Chmiela}, \citenamefont {Tkatchenko}, \citenamefont {Sauceda}, \citenamefont
  {Poltavsky}, \citenamefont {Sch{\"u}tt},\ and\ \citenamefont
  {M{\"u}ller}}]{chmiela2017machine}%
  \BibitemOpen
  \bibfield  {author} {\bibinfo {author} {\bibfnamefont {S.}~\bibnamefont
  {Chmiela}}, \bibinfo {author} {\bibfnamefont {A.}~\bibnamefont {Tkatchenko}},
  \bibinfo {author} {\bibfnamefont {H.~E.}\ \bibnamefont {Sauceda}}, \bibinfo
  {author} {\bibfnamefont {I.}~\bibnamefont {Poltavsky}}, \bibinfo {author}
  {\bibfnamefont {K.~T.}\ \bibnamefont {Sch{\"u}tt}},\ and\ \bibinfo {author}
  {\bibfnamefont {K.-R.}\ \bibnamefont {M{\"u}ller}},\ }\bibfield  {title}
  {\enquote {\bibinfo {title} {{M}achine {L}earning of {A}ccurate
  {E}nergy-{C}onserving {M}olecular {F}orce {F}ields},}\ }\href@noop {}
  {\bibfield  {journal} {\bibinfo  {journal} {Sci. Adv.}\ }\textbf {\bibinfo
  {volume} {3}},\ \bibinfo {pages} {e1603015} (\bibinfo {year}
  {2017})}\BibitemShut {NoStop}%
\bibitem [{\citenamefont {Christensen}\ \emph {et~al.}(2020)\citenamefont
  {Christensen}, \citenamefont {Bratholm}, \citenamefont {Faber},\ and\
  \citenamefont {Anatole~von Lilienfeld}}]{christensen2020fchl}%
  \BibitemOpen
  \bibfield  {author} {\bibinfo {author} {\bibfnamefont {A.~S.}\ \bibnamefont
  {Christensen}}, \bibinfo {author} {\bibfnamefont {L.~A.}\ \bibnamefont
  {Bratholm}}, \bibinfo {author} {\bibfnamefont {F.~A.}\ \bibnamefont
  {Faber}},\ and\ \bibinfo {author} {\bibfnamefont {O.}~\bibnamefont
  {Anatole~von Lilienfeld}},\ }\bibfield  {title} {\enquote {\bibinfo {title}
  {{FCHL} revisited: {F}aster and more accurate quantum machine learning},}\
  }\href@noop {} {\bibfield  {journal} {\bibinfo  {journal} {J. Chem. Phys}\
  }\textbf {\bibinfo {volume} {152}},\ \bibinfo {pages} {044107} (\bibinfo
  {year} {2020})}\BibitemShut {NoStop}%
\bibitem [{\citenamefont {Bart{\'o}k}\ \emph {et~al.}(2010)\citenamefont
  {Bart{\'o}k}, \citenamefont {Payne}, \citenamefont {Kondor},\ and\
  \citenamefont {Cs{\'a}nyi}}]{bartok2010gaussian}%
  \BibitemOpen
  \bibfield  {author} {\bibinfo {author} {\bibfnamefont {A.~P.}\ \bibnamefont
  {Bart{\'o}k}}, \bibinfo {author} {\bibfnamefont {M.~C.}\ \bibnamefont
  {Payne}}, \bibinfo {author} {\bibfnamefont {R.}~\bibnamefont {Kondor}},\ and\
  \bibinfo {author} {\bibfnamefont {G.}~\bibnamefont {Cs{\'a}nyi}},\ }\bibfield
   {title} {\enquote {\bibinfo {title} {{G}aussian approximation potentials:
  {T}he accuracy of quantum mechanics, without the electrons},}\ }\href@noop {}
  {\bibfield  {journal} {\bibinfo  {journal} {Phys. Rev. Lett.}\ }\textbf
  {\bibinfo {volume} {104}},\ \bibinfo {pages} {136403} (\bibinfo {year}
  {2010})}\BibitemShut {NoStop}%
\bibitem [{\citenamefont {Sch{\"u}tt}\ \emph {et~al.}(2017)\citenamefont
  {Sch{\"u}tt}, \citenamefont {Kindermans}, \citenamefont {Sauceda},
  \citenamefont {Chmiela}, \citenamefont {Tkatchenko},\ and\ \citenamefont
  {M{\"u}ller}}]{schutt2017schnet}%
  \BibitemOpen
  \bibfield  {author} {\bibinfo {author} {\bibfnamefont {K.}~\bibnamefont
  {Sch{\"u}tt}}, \bibinfo {author} {\bibfnamefont {P.-J.}\ \bibnamefont
  {Kindermans}}, \bibinfo {author} {\bibfnamefont {H.~E.}\ \bibnamefont
  {Sauceda}}, \bibinfo {author} {\bibfnamefont {S.}~\bibnamefont {Chmiela}},
  \bibinfo {author} {\bibfnamefont {A.}~\bibnamefont {Tkatchenko}},\ and\
  \bibinfo {author} {\bibfnamefont {K.-R.}\ \bibnamefont {M{\"u}ller}},\
  }\bibfield  {title} {\enquote {\bibinfo {title} {{S}ch{N}et: {A}
  continuous-filter convolutional neural network for modeling quantum
  interactions},}\ }\href@noop {} {\bibfield  {journal} {\bibinfo  {journal}
  {Advances in Neural Information Processing Systems}\ ,\ \bibinfo {pages}
  {991--1001}} (\bibinfo {year} {2017})}\BibitemShut {NoStop}%
\bibitem [{\citenamefont {Miller}\ \emph {et~al.}(2020)\citenamefont {Miller},
  \citenamefont {Geiger}, \citenamefont {Smidt},\ and\ \citenamefont
  {No{\'e}}}]{miller2020relevance}%
  \BibitemOpen
  \bibfield  {author} {\bibinfo {author} {\bibfnamefont {B.~K.}\ \bibnamefont
  {Miller}}, \bibinfo {author} {\bibfnamefont {M.}~\bibnamefont {Geiger}},
  \bibinfo {author} {\bibfnamefont {T.~E.}\ \bibnamefont {Smidt}},\ and\
  \bibinfo {author} {\bibfnamefont {F.}~\bibnamefont {No{\'e}}},\ }\bibfield
  {title} {\enquote {\bibinfo {title} {Relevance of rotationally equivariant
  convolutions for predicting molecular properties},}\ }\href@noop {}
  {\bibfield  {journal} {\bibinfo  {journal} {arXiv preprint arXiv:2008.08461}\
  } (\bibinfo {year} {2020})}\BibitemShut {NoStop}%
\bibitem [{\citenamefont {Klicpera}, \citenamefont {Gro{\ss}},\ and\
  \citenamefont {G{\"u}nnemann}(2020{\natexlab{a}})}]{klicpera2020directional}%
  \BibitemOpen
  \bibfield  {author} {\bibinfo {author} {\bibfnamefont {J.}~\bibnamefont
  {Klicpera}}, \bibinfo {author} {\bibfnamefont {J.}~\bibnamefont {Gro{\ss}}},\
  and\ \bibinfo {author} {\bibfnamefont {S.}~\bibnamefont {G{\"u}nnemann}},\
  }\bibfield  {title} {\enquote {\bibinfo {title} {Directional message passing
  for molecular graphs},}\ }\href@noop {} {\bibfield  {journal} {\bibinfo
  {journal} {International Conference on Learning Representations}\ } (\bibinfo
  {year} {2020}{\natexlab{a}})}\BibitemShut {NoStop}%
\bibitem [{\citenamefont {Cohen}\ and\ \citenamefont
  {Welling}(2017)}]{cohen2016steerable}%
  \BibitemOpen
  \bibfield  {author} {\bibinfo {author} {\bibfnamefont {T.~S.}\ \bibnamefont
  {Cohen}}\ and\ \bibinfo {author} {\bibfnamefont {M.}~\bibnamefont
  {Welling}},\ }\bibfield  {title} {\enquote {\bibinfo {title} {Steerable
  {CNN}s},}\ }\href@noop {} {\bibfield  {journal} {\bibinfo  {journal}
  {International Conference on Learning Representations}\ } (\bibinfo {year}
  {2017})}\BibitemShut {NoStop}%
\bibitem [{\citenamefont {Weiler}, \citenamefont {Hamprecht},\ and\
  \citenamefont {Storath}(2018)}]{weiler2018learning}%
  \BibitemOpen
  \bibfield  {author} {\bibinfo {author} {\bibfnamefont {M.}~\bibnamefont
  {Weiler}}, \bibinfo {author} {\bibfnamefont {F.~A.}\ \bibnamefont
  {Hamprecht}},\ and\ \bibinfo {author} {\bibfnamefont {M.}~\bibnamefont
  {Storath}},\ }\bibfield  {title} {\enquote {\bibinfo {title} {{L}earning
  {S}teerable {F}ilters for {R}otation {E}quivariant {CNN}s},}\ }\href@noop {}
  {\bibfield  {journal} {\bibinfo  {journal} {Proceedings of the IEEE
  Conference on Computer Vision and Pattern Recognition}\ ,\ \bibinfo {pages}
  {849--858}} (\bibinfo {year} {2018})}\BibitemShut {NoStop}%
\bibitem [{\citenamefont {Worrall}\ and\ \citenamefont
  {Brostow}(2018)}]{worrall2018cubenet}%
  \BibitemOpen
  \bibfield  {author} {\bibinfo {author} {\bibfnamefont {D.}~\bibnamefont
  {Worrall}}\ and\ \bibinfo {author} {\bibfnamefont {G.}~\bibnamefont
  {Brostow}},\ }\bibfield  {title} {\enquote {\bibinfo {title} {{C}ube{N}et:
  {E}quivariance to {3D} {R}otation and {T}ranslation},}\ }\href@noop {}
  {\bibfield  {journal} {\bibinfo  {journal} {Proceedings of the European
  Conference on Computer Vision (ECCV)}\ ,\ \bibinfo {pages} {567--584}}
  (\bibinfo {year} {2018})}\BibitemShut {NoStop}%
\bibitem [{\citenamefont {Thomas}\ \emph {et~al.}(2018)\citenamefont {Thomas},
  \citenamefont {Smidt}, \citenamefont {Kearnes}, \citenamefont {Yang},
  \citenamefont {Li}, \citenamefont {Kohlhoff},\ and\ \citenamefont
  {Riley}}]{thomas2018tensor}%
  \BibitemOpen
  \bibfield  {author} {\bibinfo {author} {\bibfnamefont {N.}~\bibnamefont
  {Thomas}}, \bibinfo {author} {\bibfnamefont {T.}~\bibnamefont {Smidt}},
  \bibinfo {author} {\bibfnamefont {S.}~\bibnamefont {Kearnes}}, \bibinfo
  {author} {\bibfnamefont {L.}~\bibnamefont {Yang}}, \bibinfo {author}
  {\bibfnamefont {L.}~\bibnamefont {Li}}, \bibinfo {author} {\bibfnamefont
  {K.}~\bibnamefont {Kohlhoff}},\ and\ \bibinfo {author} {\bibfnamefont
  {P.}~\bibnamefont {Riley}},\ }\bibfield  {title} {\enquote {\bibinfo {title}
  {{T}ensor field networks: {R}otation-and translation-equivariant neural
  networks for {3D} point clouds},}\ }\href@noop {} {\bibfield  {journal}
  {\bibinfo  {journal} {arXiv preprint arXiv:1802.08219}\ } (\bibinfo {year}
  {2018})}\BibitemShut {NoStop}%
\bibitem [{\citenamefont {Anderson}, \citenamefont {Hy},\ and\ \citenamefont
  {Kondor}(2019)}]{anderson2019cormorant}%
  \BibitemOpen
  \bibfield  {author} {\bibinfo {author} {\bibfnamefont {B.}~\bibnamefont
  {Anderson}}, \bibinfo {author} {\bibfnamefont {T.~S.}\ \bibnamefont {Hy}},\
  and\ \bibinfo {author} {\bibfnamefont {R.}~\bibnamefont {Kondor}},\
  }\bibfield  {title} {\enquote {\bibinfo {title} {{C}ormorant: {C}ovariant
  {M}olecular {N}eural {N}etworks},}\ }\href@noop {} {\bibfield  {journal}
  {\bibinfo  {journal} {Advances in Neural Information Processing Systems}\
  }\textbf {\bibinfo {volume} {32}},\ \bibinfo {pages} {14537--14546} (\bibinfo
  {year} {2019})}\BibitemShut {NoStop}%
\bibitem [{\citenamefont {Behler}\ and\ \citenamefont
  {Parrinello}(2007)}]{behler2007generalized}%
  \BibitemOpen
  \bibfield  {author} {\bibinfo {author} {\bibfnamefont {J.}~\bibnamefont
  {Behler}}\ and\ \bibinfo {author} {\bibfnamefont {M.}~\bibnamefont
  {Parrinello}},\ }\bibfield  {title} {\enquote {\bibinfo {title} {Generalized
  neural-network representation of high-dimensional potential-energy
  surfaces},}\ }\href@noop {} {\bibfield  {journal} {\bibinfo  {journal} {Phys.
  Rev. Lett.}\ }\textbf {\bibinfo {volume} {98}},\ \bibinfo {pages} {146401}
  (\bibinfo {year} {2007})}\BibitemShut {NoStop}%
\bibitem [{\citenamefont {Scarselli}\ \emph {et~al.}(2008)\citenamefont
  {Scarselli}, \citenamefont {Gori}, \citenamefont {Tsoi}, \citenamefont
  {Hagenbuchner},\ and\ \citenamefont {Monfardini}}]{scarselli2008graph}%
  \BibitemOpen
  \bibfield  {author} {\bibinfo {author} {\bibfnamefont {F.}~\bibnamefont
  {Scarselli}}, \bibinfo {author} {\bibfnamefont {M.}~\bibnamefont {Gori}},
  \bibinfo {author} {\bibfnamefont {A.~C.}\ \bibnamefont {Tsoi}}, \bibinfo
  {author} {\bibfnamefont {M.}~\bibnamefont {Hagenbuchner}},\ and\ \bibinfo
  {author} {\bibfnamefont {G.}~\bibnamefont {Monfardini}},\ }\bibfield  {title}
  {\enquote {\bibinfo {title} {The graph neural network model},}\ }\href@noop
  {} {\bibfield  {journal} {\bibinfo  {journal} {IEEE Trans. Neural Netw.}\
  }\textbf {\bibinfo {volume} {20}},\ \bibinfo {pages} {61--80} (\bibinfo
  {year} {2008})}\BibitemShut {NoStop}%
\bibitem [{\citenamefont {Duvenaud}\ \emph {et~al.}(2015)\citenamefont
  {Duvenaud}, \citenamefont {Maclaurin}, \citenamefont {Iparraguirre},
  \citenamefont {Bombarell}, \citenamefont {Hirzel}, \citenamefont
  {Aspuru-Guzik},\ and\ \citenamefont {Adams}}]{duvenaud2015convolutional}%
  \BibitemOpen
  \bibfield  {author} {\bibinfo {author} {\bibfnamefont {D.~K.}\ \bibnamefont
  {Duvenaud}}, \bibinfo {author} {\bibfnamefont {D.}~\bibnamefont {Maclaurin}},
  \bibinfo {author} {\bibfnamefont {J.}~\bibnamefont {Iparraguirre}}, \bibinfo
  {author} {\bibfnamefont {R.}~\bibnamefont {Bombarell}}, \bibinfo {author}
  {\bibfnamefont {T.}~\bibnamefont {Hirzel}}, \bibinfo {author} {\bibfnamefont
  {A.}~\bibnamefont {Aspuru-Guzik}},\ and\ \bibinfo {author} {\bibfnamefont
  {R.~P.}\ \bibnamefont {Adams}},\ }\bibfield  {title} {\enquote {\bibinfo
  {title} {Convolutional networks on graphs for learning molecular
  fingerprints},}\ }\href@noop {} {\bibfield  {journal} {\bibinfo  {journal}
  {Advances in Neural Information Processing Systems}\ }\textbf {\bibinfo
  {volume} {28}},\ \bibinfo {pages} {2224--2232} (\bibinfo {year}
  {2015})}\BibitemShut {NoStop}%
\bibitem [{\citenamefont {Kearnes}\ \emph {et~al.}(2016)\citenamefont
  {Kearnes}, \citenamefont {McCloskey}, \citenamefont {Berndl}, \citenamefont
  {Pande},\ and\ \citenamefont {Riley}}]{kearnes2016molecular}%
  \BibitemOpen
  \bibfield  {author} {\bibinfo {author} {\bibfnamefont {S.}~\bibnamefont
  {Kearnes}}, \bibinfo {author} {\bibfnamefont {K.}~\bibnamefont {McCloskey}},
  \bibinfo {author} {\bibfnamefont {M.}~\bibnamefont {Berndl}}, \bibinfo
  {author} {\bibfnamefont {V.}~\bibnamefont {Pande}},\ and\ \bibinfo {author}
  {\bibfnamefont {P.}~\bibnamefont {Riley}},\ }\bibfield  {title} {\enquote
  {\bibinfo {title} {Molecular graph convolutions: moving beyond
  fingerprints},}\ }\href@noop {} {\bibfield  {journal} {\bibinfo  {journal}
  {J. Comput. Aided Mol. Des.}\ }\textbf {\bibinfo {volume} {30}},\ \bibinfo
  {pages} {595--608} (\bibinfo {year} {2016})}\BibitemShut {NoStop}%
\bibitem [{\citenamefont {Sch{\"{u}}tt}\ \emph {et~al.}(2017)\citenamefont
  {Sch{\"{u}}tt}, \citenamefont {Arbabzadah}, \citenamefont {Chmiela},
  \citenamefont {M{\"{u}}ller},\ and\ \citenamefont
  {Tkatchenko}}]{schutt2017deep}%
  \BibitemOpen
  \bibfield  {author} {\bibinfo {author} {\bibfnamefont {K.~T.}\ \bibnamefont
  {Sch{\"{u}}tt}}, \bibinfo {author} {\bibfnamefont {F.}~\bibnamefont
  {Arbabzadah}}, \bibinfo {author} {\bibfnamefont {S.}~\bibnamefont {Chmiela}},
  \bibinfo {author} {\bibfnamefont {K.~R.}\ \bibnamefont {M{\"{u}}ller}},\ and\
  \bibinfo {author} {\bibfnamefont {A.}~\bibnamefont {Tkatchenko}},\ }\bibfield
   {title} {\enquote {\bibinfo {title} {{Quantum-chemical insights from deep
  tensor neural networks}},}\ }\href {https://doi.org/10.1038/ncomms13890}
  {\bibfield  {journal} {\bibinfo  {journal} {Nat. Commun.}\ }\textbf {\bibinfo
  {volume} {8}} (\bibinfo {year} {2017}),\ 10.1038/ncomms13890}\BibitemShut
  {NoStop}%
\bibitem [{\citenamefont {Sch{\"u}tt}\ \emph {et~al.}(2018)\citenamefont
  {Sch{\"u}tt}, \citenamefont {Sauceda}, \citenamefont {Kindermans},
  \citenamefont {Tkatchenko},\ and\ \citenamefont
  {M{\"u}ller}}]{schutt2018schnet}%
  \BibitemOpen
  \bibfield  {author} {\bibinfo {author} {\bibfnamefont {K.~T.}\ \bibnamefont
  {Sch{\"u}tt}}, \bibinfo {author} {\bibfnamefont {H.~E.}\ \bibnamefont
  {Sauceda}}, \bibinfo {author} {\bibfnamefont {P.-J.}\ \bibnamefont
  {Kindermans}}, \bibinfo {author} {\bibfnamefont {A.}~\bibnamefont
  {Tkatchenko}},\ and\ \bibinfo {author} {\bibfnamefont {K.-R.}\ \bibnamefont
  {M{\"u}ller}},\ }\bibfield  {title} {\enquote {\bibinfo {title}
  {{S}ch{N}et--{A} deep learning architecture for molecules and materials},}\
  }\href@noop {} {\bibfield  {journal} {\bibinfo  {journal} {J. Chem. Phys}\
  }\textbf {\bibinfo {volume} {148}},\ \bibinfo {pages} {241722} (\bibinfo
  {year} {2018})}\BibitemShut {NoStop}%
\bibitem [{\citenamefont {Unke}\ and\ \citenamefont
  {Meuwly}(2019)}]{unke2019physnet}%
  \BibitemOpen
  \bibfield  {author} {\bibinfo {author} {\bibfnamefont {O.~T.}\ \bibnamefont
  {Unke}}\ and\ \bibinfo {author} {\bibfnamefont {M.}~\bibnamefont {Meuwly}},\
  }\bibfield  {title} {\enquote {\bibinfo {title} {{PhysNet}: {A} {N}eural
  {N}etwork for {P}redicting {E}nergies, {F}orces, {D}ipole {M}oments, and
  {P}artial {C}harges},}\ }\href@noop {} {\bibfield  {journal} {\bibinfo
  {journal} {J. Chem. Theory Comput.}\ }\textbf {\bibinfo {volume} {15}},\
  \bibinfo {pages} {3678--3693} (\bibinfo {year} {2019})}\BibitemShut {NoStop}%
\bibitem [{\citenamefont {Lubbers}, \citenamefont {Smith},\ and\ \citenamefont
  {Barros}(2018)}]{lubbers2018hierarchical}%
  \BibitemOpen
  \bibfield  {author} {\bibinfo {author} {\bibfnamefont {N.}~\bibnamefont
  {Lubbers}}, \bibinfo {author} {\bibfnamefont {J.~S.}\ \bibnamefont {Smith}},\
  and\ \bibinfo {author} {\bibfnamefont {K.}~\bibnamefont {Barros}},\
  }\bibfield  {title} {\enquote {\bibinfo {title} {Hierarchical modeling of
  molecular energies using a deep neural network},}\ }\href@noop {} {\bibfield
  {journal} {\bibinfo  {journal} {J. Chem. Phys}\ }\textbf {\bibinfo {volume}
  {148}},\ \bibinfo {pages} {241715} (\bibinfo {year} {2018})}\BibitemShut
  {NoStop}%
\bibitem [{\citenamefont {Weiler}\ \emph {et~al.}(2018)\citenamefont {Weiler},
  \citenamefont {Geiger}, \citenamefont {Welling}, \citenamefont {Boomsma},\
  and\ \citenamefont {Cohen}}]{weiler20183d}%
  \BibitemOpen
  \bibfield  {author} {\bibinfo {author} {\bibfnamefont {M.}~\bibnamefont
  {Weiler}}, \bibinfo {author} {\bibfnamefont {M.}~\bibnamefont {Geiger}},
  \bibinfo {author} {\bibfnamefont {M.}~\bibnamefont {Welling}}, \bibinfo
  {author} {\bibfnamefont {W.}~\bibnamefont {Boomsma}},\ and\ \bibinfo {author}
  {\bibfnamefont {T.~S.}\ \bibnamefont {Cohen}},\ }\bibfield  {title} {\enquote
  {\bibinfo {title} {{3D} {S}teerable {CNN}s: {L}earning {R}otationally
  {E}quivariant {F}eatures in {V}olumetric {D}ata},}\ }\href@noop {} {\bibfield
   {journal} {\bibinfo  {journal} {Advances in Neural Information Processing
  Systems}\ }\textbf {\bibinfo {volume} {31}},\ \bibinfo {pages} {10381--10392}
  (\bibinfo {year} {2018})}\BibitemShut {NoStop}%
\bibitem [{\citenamefont {Batzner}\ \emph {et~al.}(2021)\citenamefont
  {Batzner}, \citenamefont {Smidt}, \citenamefont {Sun}, \citenamefont
  {Mailoa}, \citenamefont {Kornbluth}, \citenamefont {Molinari},\ and\
  \citenamefont {Kozinsky}}]{batzner2021se}%
  \BibitemOpen
  \bibfield  {author} {\bibinfo {author} {\bibfnamefont {S.}~\bibnamefont
  {Batzner}}, \bibinfo {author} {\bibfnamefont {T.~E.}\ \bibnamefont {Smidt}},
  \bibinfo {author} {\bibfnamefont {L.}~\bibnamefont {Sun}}, \bibinfo {author}
  {\bibfnamefont {J.~P.}\ \bibnamefont {Mailoa}}, \bibinfo {author}
  {\bibfnamefont {M.}~\bibnamefont {Kornbluth}}, \bibinfo {author}
  {\bibfnamefont {N.}~\bibnamefont {Molinari}},\ and\ \bibinfo {author}
  {\bibfnamefont {B.}~\bibnamefont {Kozinsky}},\ }\bibfield  {title} {\enquote
  {\bibinfo {title} {{SE} (3)-{E}quivariant {G}raph {N}eural {N}etworks for
  {D}ata-{E}fficient and {A}ccurate {I}nteratomic {P}otentials},}\ }\href@noop
  {} {\bibfield  {journal} {\bibinfo  {journal} {arXiv preprint
  arXiv:2101.03164}\ } (\bibinfo {year} {2021})}\BibitemShut {NoStop}%
\bibitem [{\citenamefont {Kondor}\ and\ \citenamefont
  {Trivedi}(2018)}]{kondor2018generalization}%
  \BibitemOpen
  \bibfield  {author} {\bibinfo {author} {\bibfnamefont {R.}~\bibnamefont
  {Kondor}}\ and\ \bibinfo {author} {\bibfnamefont {S.}~\bibnamefont
  {Trivedi}},\ }\bibfield  {title} {\enquote {\bibinfo {title} {On the
  {G}eneralization of {E}quivariance and {C}onvolution in {N}eural {N}etworks
  to the {A}ction of {C}ompact {G}roups},}\ }\href@noop {} {\bibfield
  {journal} {\bibinfo  {journal} {International Conference on Machine
  Learning}\ ,\ \bibinfo {pages} {2747--2755}} (\bibinfo {year}
  {2018})}\BibitemShut {NoStop}%
\bibitem [{\citenamefont {Jing}\ \emph {et~al.}(2021)\citenamefont {Jing},
  \citenamefont {Eismann}, \citenamefont {Suriana}, \citenamefont {Townshend},\
  and\ \citenamefont {Dror}}]{jing2021learning}%
  \BibitemOpen
  \bibfield  {author} {\bibinfo {author} {\bibfnamefont {B.}~\bibnamefont
  {Jing}}, \bibinfo {author} {\bibfnamefont {S.}~\bibnamefont {Eismann}},
  \bibinfo {author} {\bibfnamefont {P.}~\bibnamefont {Suriana}}, \bibinfo
  {author} {\bibfnamefont {R.~J.~L.}\ \bibnamefont {Townshend}},\ and\ \bibinfo
  {author} {\bibfnamefont {R.}~\bibnamefont {Dror}},\ }\bibfield  {title}
  {\enquote {\bibinfo {title} {Learning from protein structure with geometric
  vector perceptrons},}\ }in\ \href
  {https://openreview.net/forum?id=1YLJDvSx6J4} {\emph {\bibinfo {booktitle}
  {International Conference on Learning Representations}}}\ (\bibinfo {year}
  {2021})\BibitemShut {NoStop}%
\bibitem [{\citenamefont {Hinton}, \citenamefont {Krizhevsky},\ and\
  \citenamefont {Wang}(2011)}]{hinton2011transforming}%
  \BibitemOpen
  \bibfield  {author} {\bibinfo {author} {\bibfnamefont {G.~E.}\ \bibnamefont
  {Hinton}}, \bibinfo {author} {\bibfnamefont {A.}~\bibnamefont {Krizhevsky}},\
  and\ \bibinfo {author} {\bibfnamefont {S.~D.}\ \bibnamefont {Wang}},\
  }\bibfield  {title} {\enquote {\bibinfo {title} {Transforming
  auto-encoders},}\ }\href@noop {} {\bibfield  {journal} {\bibinfo  {journal}
  {International Conference on Artificial Neural Networks}\ ,\ \bibinfo {pages}
  {44--51}} (\bibinfo {year} {2011})}\BibitemShut {NoStop}%
\bibitem [{\citenamefont {Klicpera}, \citenamefont {Gro{\ss}},\ and\
  \citenamefont {G{\"u}nnemann}(2020{\natexlab{b}})}]{klicpera2020fast}%
  \BibitemOpen
  \bibfield  {author} {\bibinfo {author} {\bibfnamefont {J.}~\bibnamefont
  {Klicpera}}, \bibinfo {author} {\bibfnamefont {J.}~\bibnamefont {Gro{\ss}}},\
  and\ \bibinfo {author} {\bibfnamefont {S.}~\bibnamefont {G{\"u}nnemann}},\
  }\bibfield  {title} {\enquote {\bibinfo {title} {{F}ast and
  {U}ncertainty-{A}ware {D}irectional {M}essage {P}assing for
  {N}on-{E}quilibrium {M}olecules},}\ }\href@noop {} {\bibfield  {journal}
  {\bibinfo  {journal} {Machine Learning for Molecules Workshop at NeurIPS}\ }
  (\bibinfo {year} {2020}{\natexlab{b}})}\BibitemShut {NoStop}%
\bibitem [{\citenamefont {Pozdnyakov}\ \emph {et~al.}(2020)\citenamefont
  {Pozdnyakov}, \citenamefont {Willatt}, \citenamefont {Bart{\'o}k},
  \citenamefont {Ortner}, \citenamefont {Cs{\'a}nyi},\ and\ \citenamefont
  {Ceriotti}}]{pozdnyakov2020incompleteness}%
  \BibitemOpen
  \bibfield  {author} {\bibinfo {author} {\bibfnamefont {S.~N.}\ \bibnamefont
  {Pozdnyakov}}, \bibinfo {author} {\bibfnamefont {M.~J.}\ \bibnamefont
  {Willatt}}, \bibinfo {author} {\bibfnamefont {A.~P.}\ \bibnamefont
  {Bart{\'o}k}}, \bibinfo {author} {\bibfnamefont {C.}~\bibnamefont {Ortner}},
  \bibinfo {author} {\bibfnamefont {G.}~\bibnamefont {Cs{\'a}nyi}},\ and\
  \bibinfo {author} {\bibfnamefont {M.}~\bibnamefont {Ceriotti}},\ }\bibfield
  {title} {\enquote {\bibinfo {title} {Incompleteness of atomic structure
  representations},}\ }\href@noop {} {\bibfield  {journal} {\bibinfo  {journal}
  {Physical Review Letters}\ }\textbf {\bibinfo {volume} {125}},\ \bibinfo
  {pages} {166001} (\bibinfo {year} {2020})}\BibitemShut {NoStop}%
\bibitem [{\citenamefont {Hermann}\ \emph {et~al.}(2007)\citenamefont
  {Hermann}, \citenamefont {Krawczyk}, \citenamefont {Lein}, \citenamefont
  {Schwerdtfeger}, \citenamefont {Hamilton},\ and\ \citenamefont
  {Stewart}}]{hermann2007convergence}%
  \BibitemOpen
  \bibfield  {author} {\bibinfo {author} {\bibfnamefont {A.}~\bibnamefont
  {Hermann}}, \bibinfo {author} {\bibfnamefont {R.~P.}\ \bibnamefont
  {Krawczyk}}, \bibinfo {author} {\bibfnamefont {M.}~\bibnamefont {Lein}},
  \bibinfo {author} {\bibfnamefont {P.}~\bibnamefont {Schwerdtfeger}}, \bibinfo
  {author} {\bibfnamefont {I.~P.}\ \bibnamefont {Hamilton}},\ and\ \bibinfo
  {author} {\bibfnamefont {J.~J.}\ \bibnamefont {Stewart}},\ }\bibfield
  {title} {\enquote {\bibinfo {title} {Convergence of the many-body expansion
  of interaction potentials: From van der waals to covalent and metallic
  systems},}\ }\href@noop {} {\bibfield  {journal} {\bibinfo  {journal}
  {Physical Review A}\ }\textbf {\bibinfo {volume} {76}},\ \bibinfo {pages}
  {013202} (\bibinfo {year} {2007})}\BibitemShut {NoStop}%
\bibitem [{\citenamefont {Bart{\'o}k}, \citenamefont {Kondor},\ and\
  \citenamefont {Cs{\'a}nyi}(2013)}]{bartok2013representing}%
  \BibitemOpen
  \bibfield  {author} {\bibinfo {author} {\bibfnamefont {A.~P.}\ \bibnamefont
  {Bart{\'o}k}}, \bibinfo {author} {\bibfnamefont {R.}~\bibnamefont {Kondor}},\
  and\ \bibinfo {author} {\bibfnamefont {G.}~\bibnamefont {Cs{\'a}nyi}},\
  }\bibfield  {title} {\enquote {\bibinfo {title} {On representing chemical
  environments},}\ }\href@noop {} {\bibfield  {journal} {\bibinfo  {journal}
  {Phys. Rev. B}\ }\textbf {\bibinfo {volume} {87}},\ \bibinfo {pages} {184115}
  (\bibinfo {year} {2013})}\BibitemShut {NoStop}%
\bibitem [{\citenamefont {Gastegger}, \citenamefont {Sch{\"u}tt},\ and\
  \citenamefont {M{\"u}ller}(2020)}]{gastegger2020machine}%
  \BibitemOpen
  \bibfield  {author} {\bibinfo {author} {\bibfnamefont {M.}~\bibnamefont
  {Gastegger}}, \bibinfo {author} {\bibfnamefont {K.~T.}\ \bibnamefont
  {Sch{\"u}tt}},\ and\ \bibinfo {author} {\bibfnamefont {K.-R.}\ \bibnamefont
  {M{\"u}ller}},\ }\bibfield  {title} {\enquote {\bibinfo {title} {Machine
  learning of solvent effects on molecular spectra and reactions},}\
  }\href@noop {} {\bibfield  {journal} {\bibinfo  {journal} {arXiv preprint
  arXiv:2010.14942}\ } (\bibinfo {year} {2020})}\BibitemShut {NoStop}%
\bibitem [{\citenamefont {Behler}(2011)}]{behler2011atom}%
  \BibitemOpen
  \bibfield  {author} {\bibinfo {author} {\bibfnamefont {J.}~\bibnamefont
  {Behler}},\ }\bibfield  {title} {\enquote {\bibinfo {title} {Atom-centered
  symmetry functions for constructing high-dimensional neural network
  potentials},}\ }\href@noop {} {\bibfield  {journal} {\bibinfo  {journal} {J.
  Chem. Phys}\ }\textbf {\bibinfo {volume} {134}},\ \bibinfo {pages} {074106}
  (\bibinfo {year} {2011})}\BibitemShut {NoStop}%
\bibitem [{\citenamefont {Gastegger}, \citenamefont {Behler},\ and\
  \citenamefont {Marquetand}(2017)}]{gastegger2017machine}%
  \BibitemOpen
  \bibfield  {author} {\bibinfo {author} {\bibfnamefont {M.}~\bibnamefont
  {Gastegger}}, \bibinfo {author} {\bibfnamefont {J.}~\bibnamefont {Behler}},\
  and\ \bibinfo {author} {\bibfnamefont {P.}~\bibnamefont {Marquetand}},\
  }\bibfield  {title} {\enquote {\bibinfo {title} {Machine learning molecular
  dynamics for the simulation of infrared spectra},}\ }\href@noop {} {\bibfield
   {journal} {\bibinfo  {journal} {Chem. Sci.}\ }\textbf {\bibinfo {volume}
  {8}},\ \bibinfo {pages} {6924--6935} (\bibinfo {year} {2017})}\BibitemShut
  {NoStop}%
\bibitem [{\citenamefont {Veit}\ \emph {et~al.}(2020)\citenamefont {Veit},
  \citenamefont {Wilkins}, \citenamefont {Yang}, \citenamefont {DiStasio~Jr},\
  and\ \citenamefont {Ceriotti}}]{veit2020predicting}%
  \BibitemOpen
  \bibfield  {author} {\bibinfo {author} {\bibfnamefont {M.}~\bibnamefont
  {Veit}}, \bibinfo {author} {\bibfnamefont {D.~M.}\ \bibnamefont {Wilkins}},
  \bibinfo {author} {\bibfnamefont {Y.}~\bibnamefont {Yang}}, \bibinfo {author}
  {\bibfnamefont {R.~A.}\ \bibnamefont {DiStasio~Jr}},\ and\ \bibinfo {author}
  {\bibfnamefont {M.}~\bibnamefont {Ceriotti}},\ }\bibfield  {title} {\enquote
  {\bibinfo {title} {Predicting molecular dipole moments by combining atomic
  partial charges and atomic dipoles},}\ }\href@noop {} {\bibfield  {journal}
  {\bibinfo  {journal} {J. Chem. Phys.}\ }\textbf {\bibinfo {volume} {153}},\
  \bibinfo {pages} {024113} (\bibinfo {year} {2020})}\BibitemShut {NoStop}%
\bibitem [{\citenamefont {Paszke}\ \emph {et~al.}(2019)\citenamefont {Paszke},
  \citenamefont {Gross}, \citenamefont {Massa}, \citenamefont {Lerer},
  \citenamefont {Bradbury}, \citenamefont {Chanan}, \citenamefont {Killeen},
  \citenamefont {Lin}, \citenamefont {Gimelshein}, \citenamefont {Antiga} \emph
  {et~al.}}]{paszke2019pytorch}%
  \BibitemOpen
  \bibfield  {author} {\bibinfo {author} {\bibfnamefont {A.}~\bibnamefont
  {Paszke}}, \bibinfo {author} {\bibfnamefont {S.}~\bibnamefont {Gross}},
  \bibinfo {author} {\bibfnamefont {F.}~\bibnamefont {Massa}}, \bibinfo
  {author} {\bibfnamefont {A.}~\bibnamefont {Lerer}}, \bibinfo {author}
  {\bibfnamefont {J.}~\bibnamefont {Bradbury}}, \bibinfo {author}
  {\bibfnamefont {G.}~\bibnamefont {Chanan}}, \bibinfo {author} {\bibfnamefont
  {T.}~\bibnamefont {Killeen}}, \bibinfo {author} {\bibfnamefont
  {Z.}~\bibnamefont {Lin}}, \bibinfo {author} {\bibfnamefont {N.}~\bibnamefont
  {Gimelshein}}, \bibinfo {author} {\bibfnamefont {L.}~\bibnamefont {Antiga}},
  \emph {et~al.},\ }\bibfield  {title} {\enquote {\bibinfo {title} {{P}ytorch:
  {A}n imperative style, high-performance deep learning library},}\ }\href@noop
  {} {\bibfield  {journal} {\bibinfo  {journal} {arXiv preprint
  arXiv:1912.01703}\ } (\bibinfo {year} {2019})}\BibitemShut {NoStop}%
\bibitem [{\citenamefont {Sch\"utt}\ \emph {et~al.}(2018)\citenamefont
  {Sch\"utt}, \citenamefont {Kessel}, \citenamefont {Gastegger}, \citenamefont
  {Nicoli}, \citenamefont {Tkatchenko},\ and\ \citenamefont
  {M{\"u}ller}}]{schutt2018schnetpack}%
  \BibitemOpen
  \bibfield  {author} {\bibinfo {author} {\bibfnamefont {K.}~\bibnamefont
  {Sch\"utt}}, \bibinfo {author} {\bibfnamefont {P.}~\bibnamefont {Kessel}},
  \bibinfo {author} {\bibfnamefont {M.}~\bibnamefont {Gastegger}}, \bibinfo
  {author} {\bibfnamefont {K.}~\bibnamefont {Nicoli}}, \bibinfo {author}
  {\bibfnamefont {A.}~\bibnamefont {Tkatchenko}},\ and\ \bibinfo {author}
  {\bibfnamefont {K.-R.}\ \bibnamefont {M{\"u}ller}},\ }\bibfield  {title}
  {\enquote {\bibinfo {title} {{SchNetPack}: A deep learning toolbox for
  atomistic systems},}\ }\href@noop {} {\bibfield  {journal} {\bibinfo
  {journal} {J. Chem. Theory Comput.}\ }\textbf {\bibinfo {volume} {15}},\
  \bibinfo {pages} {448--455} (\bibinfo {year} {2018})}\BibitemShut {NoStop}%
\bibitem [{\citenamefont {Kingma}\ and\ \citenamefont
  {Ba}(2014)}]{kingma2014adam}%
  \BibitemOpen
  \bibfield  {author} {\bibinfo {author} {\bibfnamefont {D.~P.}\ \bibnamefont
  {Kingma}}\ and\ \bibinfo {author} {\bibfnamefont {J.}~\bibnamefont {Ba}},\
  }\bibfield  {title} {\enquote {\bibinfo {title} {{A}dam: {A} method for
  {S}tochastic {O}ptimization},}\ }\href@noop {} {\bibfield  {journal}
  {\bibinfo  {journal} {arXiv preprint arXiv:1412.6980}\ } (\bibinfo {year}
  {2014})}\BibitemShut {NoStop}%
\bibitem [{\citenamefont {Ramakrishnan}\ \emph {et~al.}(2014)\citenamefont
  {Ramakrishnan}, \citenamefont {Dral}, \citenamefont {Rupp},\ and\
  \citenamefont {Von~Lilienfeld}}]{ramakrishnan2014quantum}%
  \BibitemOpen
  \bibfield  {author} {\bibinfo {author} {\bibfnamefont {R.}~\bibnamefont
  {Ramakrishnan}}, \bibinfo {author} {\bibfnamefont {P.~O.}\ \bibnamefont
  {Dral}}, \bibinfo {author} {\bibfnamefont {M.}~\bibnamefont {Rupp}},\ and\
  \bibinfo {author} {\bibfnamefont {O.~A.}\ \bibnamefont {Von~Lilienfeld}},\
  }\bibfield  {title} {\enquote {\bibinfo {title} {Quantum chemistry structures
  and properties of 134 kilo molecules},}\ }\href@noop {} {\bibfield  {journal}
  {\bibinfo  {journal} {Sci. Data}\ }\textbf {\bibinfo {volume} {1}},\ \bibinfo
  {pages} {1--7} (\bibinfo {year} {2014})}\BibitemShut {NoStop}%
\bibitem [{Note1()}]{Note1}%
  \BibitemOpen
  \bibinfo {note} {\protect \url
  {https://github.com/klicperajo/dimenet}}\BibitemShut {NoStop}%
\bibitem [{\citenamefont {Christensen}\ and\ \citenamefont {von
  Lilienfeld}(2020)}]{christensen2020role}%
  \BibitemOpen
  \bibfield  {author} {\bibinfo {author} {\bibfnamefont {A.~S.}\ \bibnamefont
  {Christensen}}\ and\ \bibinfo {author} {\bibfnamefont {O.~A.}\ \bibnamefont
  {von Lilienfeld}},\ }\bibfield  {title} {\enquote {\bibinfo {title} {On the
  role of gradients for machine learning of molecular energies and forces},}\
  }\href@noop {} {\bibfield  {journal} {\bibinfo  {journal} {Mach. Learn.: Sci.
  Technol.}\ }\textbf {\bibinfo {volume} {1}},\ \bibinfo {pages} {045018}
  (\bibinfo {year} {2020})}\BibitemShut {NoStop}%
\bibitem [{\citenamefont {Thomas}\ \emph {et~al.}(2013)\citenamefont {Thomas},
  \citenamefont {Brehm}, \citenamefont {Fligg}, \citenamefont {V{\"o}hringer},\
  and\ \citenamefont {Kirchner}}]{thomas2013computing}%
  \BibitemOpen
  \bibfield  {author} {\bibinfo {author} {\bibfnamefont {M.}~\bibnamefont
  {Thomas}}, \bibinfo {author} {\bibfnamefont {M.}~\bibnamefont {Brehm}},
  \bibinfo {author} {\bibfnamefont {R.}~\bibnamefont {Fligg}}, \bibinfo
  {author} {\bibfnamefont {P.}~\bibnamefont {V{\"o}hringer}},\ and\ \bibinfo
  {author} {\bibfnamefont {B.}~\bibnamefont {Kirchner}},\ }\bibfield  {title}
  {\enquote {\bibinfo {title} {Computing vibrational spectra from ab initio
  molecular dynamics},}\ }\href@noop {} {\bibfield  {journal} {\bibinfo
  {journal} {Phys. Chem. Chem. Phys.}\ }\textbf {\bibinfo {volume} {15}},\
  \bibinfo {pages} {6608--6622} (\bibinfo {year} {2013})}\BibitemShut {NoStop}%
\bibitem [{\citenamefont {Sauceda}\ \emph {et~al.}(2021)\citenamefont
  {Sauceda}, \citenamefont {Vassilev-Galindo}, \citenamefont {Chmiela},
  \citenamefont {M{\"u}ller},\ and\ \citenamefont
  {Tkatchenko}}]{sauceda2021dynamical}%
  \BibitemOpen
  \bibfield  {author} {\bibinfo {author} {\bibfnamefont {H.~E.}\ \bibnamefont
  {Sauceda}}, \bibinfo {author} {\bibfnamefont {V.}~\bibnamefont
  {Vassilev-Galindo}}, \bibinfo {author} {\bibfnamefont {S.}~\bibnamefont
  {Chmiela}}, \bibinfo {author} {\bibfnamefont {K.-R.}\ \bibnamefont
  {M{\"u}ller}},\ and\ \bibinfo {author} {\bibfnamefont {A.}~\bibnamefont
  {Tkatchenko}},\ }\bibfield  {title} {\enquote {\bibinfo {title} {Dynamical
  strengthening of covalent and non-covalent molecular interactions by nuclear
  quantum effects at finite temperature},}\ }\href@noop {} {\bibfield
  {journal} {\bibinfo  {journal} {Nat. Commun.}\ }\textbf {\bibinfo {volume}
  {12}},\ \bibinfo {pages} {1--10} (\bibinfo {year} {2021})}\BibitemShut
  {NoStop}%
\bibitem [{\citenamefont {Linstrom}\ and\ \citenamefont
  {Mallard}(2020)}]{NIST}%
  \BibitemOpen
  \bibfield  {author} {\bibinfo {author} {\bibfnamefont {P.}~\bibnamefont
  {Linstrom}}\ and\ \bibinfo {author} {\bibfnamefont {W.~G.}\ \bibnamefont
  {Mallard}},\ }\href@noop {} {\emph {\bibinfo {title} {{NIST} {C}hemistry
  {W}eb{B}ook {NIST} {S}tandard {R}eference {D}atabase {N}umber 69}}}\
  (\bibinfo  {publisher} {National Institute of Standards and Technology},\
  \bibinfo {address} {doi:10.18434/T4D303, (retrieved September 24, 2020),
  Gaithersburg MD, 20899},\ \bibinfo {year} {2020})\BibitemShut {NoStop}%
\bibitem [{\citenamefont {Kiefer}(2017)}]{kiefer2017simultaneous}%
  \BibitemOpen
  \bibfield  {author} {\bibinfo {author} {\bibfnamefont {J.}~\bibnamefont
  {Kiefer}},\ }\bibfield  {title} {\enquote {\bibinfo {title} {Simultaneous
  acquisition of the polarized and depolarized {R}aman signal with a single
  detector},}\ }\href@noop {} {\bibfield  {journal} {\bibinfo  {journal} {Anal.
  Chem.}\ }\textbf {\bibinfo {volume} {89}},\ \bibinfo {pages} {5725--5728}
  (\bibinfo {year} {2017})}\BibitemShut {NoStop}%
\bibitem [{\citenamefont {Gebauer}, \citenamefont {Gastegger},\ and\
  \citenamefont {Sch{\"u}tt}(2019)}]{gebauer2019symmetry}%
  \BibitemOpen
  \bibfield  {author} {\bibinfo {author} {\bibfnamefont {N.}~\bibnamefont
  {Gebauer}}, \bibinfo {author} {\bibfnamefont {M.}~\bibnamefont {Gastegger}},\
  and\ \bibinfo {author} {\bibfnamefont {K.}~\bibnamefont {Sch{\"u}tt}},\
  }\bibfield  {title} {\enquote {\bibinfo {title} {Symmetry-adapted generation
  of 3d point sets for the targeted discovery of molecules},}\ }\href@noop {}
  {\bibfield  {journal} {\bibinfo  {journal} {Advances in Neural Information
  Processing Systems}\ ,\ \bibinfo {pages} {7564--7576}} (\bibinfo {year}
  {2019})}\BibitemShut {NoStop}%
\bibitem [{\citenamefont {K{\"o}hler}, \citenamefont {Klein},\ and\
  \citenamefont {No{\'e}}(2019)}]{kohler2019equivariant}%
  \BibitemOpen
  \bibfield  {author} {\bibinfo {author} {\bibfnamefont {J.}~\bibnamefont
  {K{\"o}hler}}, \bibinfo {author} {\bibfnamefont {L.}~\bibnamefont {Klein}},\
  and\ \bibinfo {author} {\bibfnamefont {F.}~\bibnamefont {No{\'e}}},\
  }\bibfield  {title} {\enquote {\bibinfo {title} {{E}quivariant {F}lows:
  sampling configurations for multi-body systems with symmetric energies},}\
  }\href@noop {} {\bibfield  {journal} {\bibinfo  {journal} {Proceedings of the
  37th International Conference on Machine Learning}\ } (\bibinfo {year}
  {2019})}\BibitemShut {NoStop}%
\bibitem [{\citenamefont {Simm}\ \emph {et~al.}(2020)\citenamefont {Simm},
  \citenamefont {Pinsler}, \citenamefont {Cs{\'a}nyi},\ and\ \citenamefont
  {Hern{\'a}ndez-Lobato}}]{simm2020symmetry}%
  \BibitemOpen
  \bibfield  {author} {\bibinfo {author} {\bibfnamefont {G.~N.}\ \bibnamefont
  {Simm}}, \bibinfo {author} {\bibfnamefont {R.}~\bibnamefont {Pinsler}},
  \bibinfo {author} {\bibfnamefont {G.}~\bibnamefont {Cs{\'a}nyi}},\ and\
  \bibinfo {author} {\bibfnamefont {J.~M.}\ \bibnamefont
  {Hern{\'a}ndez-Lobato}},\ }\bibfield  {title} {\enquote {\bibinfo {title}
  {{S}ymmetry-{A}ware {A}ctor-{C}ritic for {3D} {M}olecular {D}esign},}\
  }\href@noop {} {\bibfield  {journal} {\bibinfo  {journal} {arXiv preprint
  arXiv:2011.12747}\ } (\bibinfo {year} {2020})}\BibitemShut {NoStop}%
\bibitem [{\citenamefont {Hegde}\ and\ \citenamefont
  {Bowen}(2017)}]{hegde2017machine}%
  \BibitemOpen
  \bibfield  {author} {\bibinfo {author} {\bibfnamefont {G.}~\bibnamefont
  {Hegde}}\ and\ \bibinfo {author} {\bibfnamefont {R.~C.}\ \bibnamefont
  {Bowen}},\ }\bibfield  {title} {\enquote {\bibinfo {title} {{M}achine-learned
  approximations to {D}ensity {F}unctional {T}heory {H}amiltonians},}\
  }\href@noop {} {\bibfield  {journal} {\bibinfo  {journal} {Sci. Rep.}\
  }\textbf {\bibinfo {volume} {7}},\ \bibinfo {pages} {1--11} (\bibinfo {year}
  {2017})}\BibitemShut {NoStop}%
\bibitem [{\citenamefont {Sch{\"u}tt}\ \emph {et~al.}(2019)\citenamefont
  {Sch{\"u}tt}, \citenamefont {Gastegger}, \citenamefont {Tkatchenko},
  \citenamefont {M{\"u}ller},\ and\ \citenamefont
  {Maurer}}]{schutt2019unifying}%
  \BibitemOpen
  \bibfield  {author} {\bibinfo {author} {\bibfnamefont {K.}~\bibnamefont
  {Sch{\"u}tt}}, \bibinfo {author} {\bibfnamefont {M.}~\bibnamefont
  {Gastegger}}, \bibinfo {author} {\bibfnamefont {A.}~\bibnamefont
  {Tkatchenko}}, \bibinfo {author} {\bibfnamefont {K.-R.}\ \bibnamefont
  {M{\"u}ller}},\ and\ \bibinfo {author} {\bibfnamefont {R.~J.}\ \bibnamefont
  {Maurer}},\ }\bibfield  {title} {\enquote {\bibinfo {title} {Unifying machine
  learning and quantum chemistry with a deep neural network for molecular
  wavefunctions},}\ }\href@noop {} {\bibfield  {journal} {\bibinfo  {journal}
  {Nat. Commun.}\ }\textbf {\bibinfo {volume} {10}},\ \bibinfo {pages} {1--10}
  (\bibinfo {year} {2019})}\BibitemShut {NoStop}%
\bibitem [{\citenamefont {Hermann}, \citenamefont {Sch{\"a}tzle},\ and\
  \citenamefont {No{\'e}}(2020)}]{hermann2020deep}%
  \BibitemOpen
  \bibfield  {author} {\bibinfo {author} {\bibfnamefont {J.}~\bibnamefont
  {Hermann}}, \bibinfo {author} {\bibfnamefont {Z.}~\bibnamefont
  {Sch{\"a}tzle}},\ and\ \bibinfo {author} {\bibfnamefont {F.}~\bibnamefont
  {No{\'e}}},\ }\bibfield  {title} {\enquote {\bibinfo {title}
  {{D}eep-neural-network solution of the electronic {S}chr{\"o}dinger
  equation},}\ }\href@noop {} {\bibfield  {journal} {\bibinfo  {journal} {Nat.
  Chem.}\ }\textbf {\bibinfo {volume} {12}},\ \bibinfo {pages} {891--897}
  (\bibinfo {year} {2020})}\BibitemShut {NoStop}%
\bibitem [{\citenamefont {Goyal}\ and\ \citenamefont
  {Ferrara}(2018)}]{goyal2018graph}%
  \BibitemOpen
  \bibfield  {author} {\bibinfo {author} {\bibfnamefont {P.}~\bibnamefont
  {Goyal}}\ and\ \bibinfo {author} {\bibfnamefont {E.}~\bibnamefont
  {Ferrara}},\ }\bibfield  {title} {\enquote {\bibinfo {title} {Graph embedding
  techniques, applications, and performance: A survey},}\ }\href@noop {}
  {\bibfield  {journal} {\bibinfo  {journal} {Knowledge-Based Systems}\
  }\textbf {\bibinfo {volume} {151}},\ \bibinfo {pages} {78--94} (\bibinfo
  {year} {2018})}\BibitemShut {NoStop}%
\bibitem [{\citenamefont {Bannwarth}, \citenamefont {Ehlert},\ and\
  \citenamefont {Grimme}(2019)}]{bannwarth2019gfn2}%
  \BibitemOpen
  \bibfield  {author} {\bibinfo {author} {\bibfnamefont {C.}~\bibnamefont
  {Bannwarth}}, \bibinfo {author} {\bibfnamefont {S.}~\bibnamefont {Ehlert}},\
  and\ \bibinfo {author} {\bibfnamefont {S.}~\bibnamefont {Grimme}},\
  }\bibfield  {title} {\enquote {\bibinfo {title} {{GFN2}-x{TB} -- {A}n
  accurate and broadly parametrized self-consistent tight-binding quantum
  chemical method with multipole electrostatics and density-dependent
  dispersion contributions},}\ }\href@noop {} {\bibfield  {journal} {\bibinfo
  {journal} {J. Chem. Theory Comput.}\ }\textbf {\bibinfo {volume} {15}},\
  \bibinfo {pages} {1652--1671} (\bibinfo {year} {2019})}\BibitemShut {NoStop}%
\bibitem [{Note2()}]{Note2}%
  \BibitemOpen
  \bibinfo {note} {\protect \url
  {https://github.com/grimme-lab/xtb}}\BibitemShut {NoStop}%
\bibitem [{\citenamefont {Smith}, \citenamefont {Isayev},\ and\ \citenamefont
  {Roitberg}(2017)}]{smith2017ani}%
  \BibitemOpen
  \bibfield  {author} {\bibinfo {author} {\bibfnamefont {J.~S.}\ \bibnamefont
  {Smith}}, \bibinfo {author} {\bibfnamefont {O.}~\bibnamefont {Isayev}},\ and\
  \bibinfo {author} {\bibfnamefont {A.~E.}\ \bibnamefont {Roitberg}},\
  }\bibfield  {title} {\enquote {\bibinfo {title} {{ANI-1}: an extensible
  neural network potential with {DFT} accuracy at force field computational
  cost},}\ }\href@noop {} {\bibfield  {journal} {\bibinfo  {journal} {Chem.
  Sci.}\ }\textbf {\bibinfo {volume} {8}},\ \bibinfo {pages} {3192--3203}
  (\bibinfo {year} {2017})}\BibitemShut {NoStop}%
\bibitem [{\citenamefont {Adamo}\ and\ \citenamefont
  {Barone}(1999)}]{adamo1999toward}%
  \BibitemOpen
  \bibfield  {author} {\bibinfo {author} {\bibfnamefont {C.}~\bibnamefont
  {Adamo}}\ and\ \bibinfo {author} {\bibfnamefont {V.}~\bibnamefont {Barone}},\
  }\bibfield  {title} {\enquote {\bibinfo {title} {Toward reliable density
  functional methods without adjustable parameters: The {PBE0} model},}\
  }\href@noop {} {\bibfield  {journal} {\bibinfo  {journal} {J. Chem. Phys.}\
  }\textbf {\bibinfo {volume} {110}},\ \bibinfo {pages} {6158--6170} (\bibinfo
  {year} {1999})}\BibitemShut {NoStop}%
\bibitem [{\citenamefont {Weigend}\ and\ \citenamefont
  {Ahlrichs}(2005)}]{Weigend2005PCCP}%
  \BibitemOpen
  \bibfield  {author} {\bibinfo {author} {\bibfnamefont {F.}~\bibnamefont
  {Weigend}}\ and\ \bibinfo {author} {\bibfnamefont {R.}~\bibnamefont
  {Ahlrichs}},\ }\bibfield  {title} {\enquote {\bibinfo {title} {Balanced basis
  sets of split valence, triple zeta valence and quadruple zeta valence quality
  for {H} to {Rn}: {D}esign and assessment of accuracy},}\ }\href
  {https://doi.org/10.1039/B508541A} {\bibfield  {journal} {\bibinfo  {journal}
  {Phys. Chem. Chem. Phys.}\ }\textbf {\bibinfo {volume} {7}},\ \bibinfo
  {pages} {3297--3305} (\bibinfo {year} {2005})}\BibitemShut {NoStop}%
\bibitem [{\citenamefont {Neese}(2012)}]{Neese2012WCMS}%
  \BibitemOpen
  \bibfield  {author} {\bibinfo {author} {\bibfnamefont {F.}~\bibnamefont
  {Neese}},\ }\bibfield  {title} {\enquote {\bibinfo {title} {The {ORCA}
  program system},}\ }\href {https://doi.org/10.1002/wcms.81} {\bibfield
  {journal} {\bibinfo  {journal} {WIREs Comput. Mol. Sci.}\ }\textbf {\bibinfo
  {volume} {2}},\ \bibinfo {pages} {73--78} (\bibinfo {year}
  {2012})}\BibitemShut {NoStop}%
\bibitem [{\citenamefont {Weigend}(2002)}]{weigend2002fully}%
  \BibitemOpen
  \bibfield  {author} {\bibinfo {author} {\bibfnamefont {F.}~\bibnamefont
  {Weigend}},\ }\bibfield  {title} {\enquote {\bibinfo {title} {{A} fully
  direct {RI-HF} algorithm: {I}mplementation, optimised auxiliary basis sets,
  demonstration of accuracy and efficiency},}\ }\href@noop {} {\bibfield
  {journal} {\bibinfo  {journal} {Phys. Chem. Chem. Phys.}\ }\textbf {\bibinfo
  {volume} {4}},\ \bibinfo {pages} {4285--4291} (\bibinfo {year}
  {2002})}\BibitemShut {NoStop}%
\bibitem [{\citenamefont {Martyna}, \citenamefont {Klein},\ and\ \citenamefont
  {Tuckerman}(1992)}]{martyna1992nhc}%
  \BibitemOpen
  \bibfield  {author} {\bibinfo {author} {\bibfnamefont {G.~J.}\ \bibnamefont
  {Martyna}}, \bibinfo {author} {\bibfnamefont {M.~L.}\ \bibnamefont {Klein}},\
  and\ \bibinfo {author} {\bibfnamefont {M.}~\bibnamefont {Tuckerman}},\
  }\bibfield  {title} {\enquote {\bibinfo {title} {{N}os{\'e}–{H}oover
  chains: {T}he canonical ensemble via continuous dynamics},}\ }\href
  {https://doi.org/10.1063/1.463940} {\bibfield  {journal} {\bibinfo  {journal}
  {J. Chem. Phys.}\ }\textbf {\bibinfo {volume} {97}},\ \bibinfo {pages}
  {2635--2643} (\bibinfo {year} {1992})}\BibitemShut {NoStop}%
\bibitem [{\citenamefont {Ceriotti}\ \emph {et~al.}(2010)\citenamefont
  {Ceriotti}, \citenamefont {Parrinello}, \citenamefont {Markland},\ and\
  \citenamefont {Manolopoulos}}]{ceriotti2010stochastic}%
  \BibitemOpen
  \bibfield  {author} {\bibinfo {author} {\bibfnamefont {M.}~\bibnamefont
  {Ceriotti}}, \bibinfo {author} {\bibfnamefont {M.}~\bibnamefont
  {Parrinello}}, \bibinfo {author} {\bibfnamefont {T.~E.}\ \bibnamefont
  {Markland}},\ and\ \bibinfo {author} {\bibfnamefont {D.~E.}\ \bibnamefont
  {Manolopoulos}},\ }\bibfield  {title} {\enquote {\bibinfo {title} {Efficient
  stochastic thermostatting of path integral molecular dynamics},}\ }\href
  {https://doi.org/10.1063/1.3489925} {\bibfield  {journal} {\bibinfo
  {journal} {J. Chem. Phys.}\ }\textbf {\bibinfo {volume} {133}},\ \bibinfo
  {pages} {124104} (\bibinfo {year} {2010})}\BibitemShut {NoStop}%
\bibitem [{\citenamefont {{W}iener}(1930)}]{wiener1930}%
  \BibitemOpen
  \bibfield  {author} {\bibinfo {author} {\bibfnamefont {N.}~\bibnamefont
  {{W}iener}},\ }\bibfield  {title} {\enquote {\bibinfo {title} {{G}eneralized
  harmonic analysis},}\ }\href {https://doi.org/10.1007/BF02546511} {\bibfield
  {journal} {\bibinfo  {journal} {Acta Math.}\ }\textbf {\bibinfo {volume}
  {55}},\ \bibinfo {pages} {117--258} (\bibinfo {year} {1930})}\BibitemShut
  {NoStop}%
\bibitem [{\citenamefont {{B}lackman}\ and\ \citenamefont
  {{T}ukey}(1958)}]{6768513}%
  \BibitemOpen
  \bibfield  {author} {\bibinfo {author} {\bibfnamefont {R.~B.}\ \bibnamefont
  {{B}lackman}}\ and\ \bibinfo {author} {\bibfnamefont {J.~W.}\ \bibnamefont
  {{T}ukey}},\ }\bibfield  {title} {\enquote {\bibinfo {title} {{T}he
  measurement of power spectra from the point of view of communications
  engineering -- {P}art {I}},}\ }\href
  {https://doi.org/10.1002/j.1538-7305.1958.tb03874.x} {\bibfield  {journal}
  {\bibinfo  {journal} {Bell Syst. Tech. J.}\ }\textbf {\bibinfo {volume}
  {37}},\ \bibinfo {pages} {185--282} (\bibinfo {year} {1958})}\BibitemShut
  {NoStop}%
\end{thebibliography}%

\renewcommand{\thefigure}{\arabic{figure}}
\renewcommand{\thetable}{\arabic{table}}
\renewcommand{\figurename}{Supplementary Figure}
\renewcommand{\tablename}{Supplementary Table}
\captionsetup[figure]{justification=raggedright}
\captionsetup[table]{justification=raggedright}
\clearpage
\appendix

\section{Data}

\subsection{Reference data for substituted ferrocene}

Reference computations for substituted ferrocene were carried out with the semi-empirical GFN2-xTB method\cite{bannwarth2019gfn2} using the \texttt{xtb} package\footnote{\url{https://github.com/grimme-lab/xtb}}). Training structures were generated by normal mode sampling\cite{smith2017ani} at 300~K starting from the global minimum structure and randomly rotating the cylopentadienyl (Cp) moieties relative to each other around the Cp-Fe-Cp axis by angles uniformly sampled from $[0,2\pi]$. The rotational potential energy profile was generated by sampling a full rotation around the Cp-Fe-Cp axis using 1k steps, starting from the global minimum structure and keeping all other degrees of freedom fixed at their equilibrium positions.

\subsection{Reference data for infrared and Raman spectra}

Electronic structure reference computations for aspirin were carried out at the PBE0/def2-TZVP\cite{adamo1999toward,Weigend2005PCCP} level of theory using the ORCA quantum chemistry package~\cite{Neese2012WCMS}.
SCF convergence was set to tight and integration grid levels of 4 and 5 were employed during SCF iterations and the final computation of properties, respectively.
Computations were accelerated using the RIJK approximation~\cite{weigend2002fully}.
The reference data for aspirin was generated by selecting 20\,000 random configurations from the MD17 database~\cite{chmiela2017machine} and recomputing them at the above level of theory.
%
% Reference data ethanol
Reference data for the ethanol molecule was taken from reference~\cite{gastegger2020machine}, which employed the same level of theory bar the RIJK approximation.

\section{Training details}
All models were trained using the Adam optimizer~\cite{kingma2014adam}.
The learning rate is decayed by a factor of 0.5 if the validation loss plateaus, starting with the largest learning rate that does not diverge in steps 1e-4, 5e-4, 1e-3, etc.
We apply exponential smoothing with factor $0.9$ to the validation loss to reduce the impact of fluctuations which are particularly common when training with both energies and forces.
We use smaller batches for datasets where we train on both energies and forces, since it is commonly observed that larger batch sizes converge to larger errors in this setting.

\begin{table*}[htb]
\caption{Training parameters for all experiments.}
\label{tab:md17}
\begin{center}
\begin{scriptsize}
\begin{sc}
\begin{tabular}{lrrrrr}
\toprule
Data set & batch size & learning rate & decay patience & stopping patience & $r_\text{cut}$ [\AA]\\ 
\midrule
\textsc{QM9} & 100 & $5 \cdot 10^{-4}$ & 5 & 30 & 5.0 \\
\textsc{MD17} & 10 & $1 \cdot 10^{-3}$ & 50 & 150 & 5.0  \\
\textsc{Ferrocene} & 10 & $1 \cdot 10^{-3}$ & 10 & 30 & 2.5-4.0 \\
\textsc{Spectra} & 10 & $5 \cdot 10^{-4}$ & 15 & 50 & 2.7 \\
\bottomrule
\end{tabular}
\end{sc}
\end{scriptsize}
\end{center}
\vskip -0.1in
\end{table*}

\section{Computation of infrared and Raman spectra}

All simulations were carried out with the molecular dynamics module implemented in SchNetPack~\cite{schutt2018schnet}.
Classical molecular dynamics simulations for ethanol and aspirin were carried out for 50~ps at a temperature of 300~K controlled via Nose-Hoover chain~\cite{martyna1992nhc} thermostat with a chain length of 3 and time constant of 100~fs. The first 10~ps of these trajectories were then discarded.
Ring polymer molecular dynamics using 64 replicas were performed using the same simulation and equilibration periods and temperature, but employed a specially adapted global Nose-Hoover chain as introduced in Ref.~\cite{ceriotti2010stochastic} for thermostatting instead.
The overall chain settings were kept the same as above.
In all cases, the velocity Verlet algorithm and a time step of 0.2~fs were used to integrate the equations of motion.
Infrared and Raman spectra were computed from the time-autocorrelation functions of the dipole moment and polarizability time derivatives (see Ref.~\cite{thomas2013computing}).
All autocorrelation functions were computed with the Wiener-Khinchin theorem~\cite{wiener1930}, using an autocorrelation depth of 2048~fs.
A Hann window function~\cite{6768513} and zero-padding were applied to the autocorrelation functions in order to enhance the spectra.
Raman spectra were calculated using a laser frequency of 514~nm and temperature of 300~K.

\end{document}